\documentclass{article}

\PassOptionsToPackage{numbers, compress}{natbib}
\usepackage[dvipsnames]{xcolor}

\definecolor{myblue}{RGB}{31, 119, 180}
\definecolor{myorange}{RGB}{255, 127, 14}
\definecolor{mygreen}{RGB}{44, 160, 44}
\definecolor{myred}{RGB}{214, 39, 40}
\definecolor{myyellow}{RGB}{230,194,0}
\usepackage[nonatbib,final]{neurips_2024}

\usepackage{natbib}
\usepackage[utf8]{inputenc} %
\usepackage[T1]{fontenc}    %
\usepackage[allcolors=NavyBlue,colorlinks=true,backref=page]{hyperref}       %
\usepackage{url}            %
\usepackage{booktabs}       %
\usepackage{amsfonts}       %
\usepackage{nicefrac}       %
\usepackage{microtype}      %
\usepackage{xspace}
\usepackage{subcaption}

\usepackage{wrapfig}

\usepackage[noend]{algpseudocode}
\usepackage{algorithm}

\usepackage{cite}
\usepackage{comment}
\usepackage{prettyref}
\usepackage{amsfonts}
\usepackage{makecell}
\usepackage{adjustbox}
\usepackage{multicol}
\usepackage{amsmath}
\usepackage{amssymb}
\usepackage{mathtools}
\usepackage{amsthm}
\usepackage{enumitem}

\usepackage[textsize=footnotesize]{todonotes}
\usepackage{xspace}

\definecolor{colorcomment}{RGB}{160, 190, 210}%
\makeatletter
\algnewcommand{\LineComment}[1]{\Statex \hskip\ALG@thistlm \(\triangleright\) 
{\color{colorcomment}#1}}
\makeatother

\makeatletter
\algnewcommand{\IndentLineComment}[1]{\Statex \hskip\ALG@tlm \(\triangleright\) {\color{colorcomment}#1}}
\makeatother

\newcommand\policy{\ensuremath{\pi}}

\newcommand\len{\mathcal{L}}
\newcommand\prob{{p}}

\newcommand\horizon{\ensuremath{T}}

\newcommand\tokenSeq{\ensuremath{\mathbf{X}}}
\newcommand\token{\ensuremath{{x}}}

\newcommand\corpus{\mathcal{C}}

\def \argmax {\mathop{\rm arg\,max}}

\newcommand\Vocabulary{\ensuremath{\mathcal{V}}}
\newcommand\maskedTokens{\ensuremath{\mathbf{M}}}
\newcommand\unmaskedTokens{\ensuremath{\mathbf{S}}}

\newcommand\likelihood{\ensuremath{\mathcal{L}}}

\newcommand{\hypothesis}{\mathcal{Y}}

\newcommand{\algname}{\textsc{ControllableGPT}\xspace}

\newif\iffinal
\finaltrue

\iffinal
    \newcommand{\fix}[1]{#1}
    \newcommand{\note}[1]{}
    \newcommand{\pref}[1]{}
    \newcommand{\XL}[1]{}
    \newcommand{\SJ}[1]{}
    \newcommand{\XLinline}[1]{}
    \newcommand{\SJinline}[1]{}
\else
    \newcommand{\fix}[1]{{\color{red} #1}}
    \newcommand{\SJ}[1]{\todo[fancyline,color=Green!40]{SJ: #1}\xspace}
    
    \newcommand{\XL}[1]{\todo[fancyline,color=Maroon!40]{XL: #1}\xspace}
    \newcommand{\XLinline}[1]{\textcolor{Maroon}{[XL: #1]}}
    \newcommand{\SJinline}[1]{\textcolor{Green}{[SJ: #1]}}
    
    \newcommand{\note}[1]{{\color{purple}[XL: #1]}}
    
    \newcommand{\pref}[1]{{\color{blue}(\ref{#1})}}
\fi

\newcommand{\tabref}[1]{Table~\ref{#1}}
\newcommand{\figref}[1]{Fig.~\ref{#1}}

\newcommand{\appref}[1]{Appendix~\ref{#1}}

\newcommand{\paren} [1] {\ensuremath{ \left( {#1} \right) }}

\newcommand{\bracket}[1]{\left[#1\right]}
\newcommand{\tuple}[1]{\ensuremath{\left\langle #1 \right\rangle}}
\newcommand{\curlybracket}[1]{\ensuremath{\left\{#1\right\}}}

\theoremstyle{plain}

\theoremstyle{definition}

\theoremstyle{remark}

\author{%
}

\title{
\algname: A Ground-Up Designed Controllable GPT for Molecule Optimization
}

\author{Xuefeng Liu\textsuperscript{1}\thanks{Correspondence to: Xuefeng Liu <\href{mailto:xuefeng@uchicago.edu}{xuefeng@uchicago.edu}>.} ,~\textbf{Songhao Jiang\textsuperscript{1}},~\textbf{Bo Li\textsuperscript{1}},~\textbf{Rick  Stevens\textsuperscript{1,2}} \\
\textsuperscript{1}Department of Computer Science, University of Chicago\\
\textsuperscript{2}Argonne National Laboratory 
}

\begin{document}

\maketitle

\begin{abstract}

Large Language Models (LLMs) employ three popular training approaches: Masked Language Models (MLM), Causal Language Models (CLM), and Sequence-to-Sequence Models (seq2seq). However, each approach has its strengths and limitations, and faces challenges in addressing specific tasks that require controllable and bidirectional generation, such as drug optimization.
To address this challenge, inspired by the biological processes of growth and evolution, which involve the expansion, shrinking, and mutation of sequences, we introduce \algname. This initiative represents the first effort to combine the advantages of MLM, CLM, and seq2seq into a single unified, controllable GPT framework. It enables the precise management of specific locations and ranges within a sequence, allowing for expansion, reduction, or mutation over chosen or random lengths, while maintaining the integrity of any specified positions or subsequences.
In this work, we designed \algname for drug optimization from the ground up, which included {proposing the Causally Masked Seq2seq (CMS) objective}, developing the training corpus, introducing a novel pre-training approach, and devising a unique generation process.
We demonstrate the effectiveness and controllability of \algname by conducting  experiments on drug optimization tasks for both viral and cancer benchmarks, surpassing competing baselines.

\end{abstract}

\section{Introduction}\label{sec:intro}

The Generative Pre-trained Transformer (GPT)~\citep{floridi2020gpt,yenduri2023generative} has achieved significant success in applications such as ChatGPT~\citep{ouyang2022training,wu2023brief} for chatbox, Copilot~\citep{barke2023grounded} for code generation and VideoGPT~\citep{yan2021videogpt} for video creation. However, GPT operates unidirectionally and lacks controllability~\citep{ethayarajh2019contextual}, which means it still faces the challenge in handling the real-world scenarios that require both controllable and bidirectional generation capabilities, as dictated by the current design of GPT.

On the other hand, drug discovery~\citep{berdigaliyev2020overview} has become increasingly important since the advent of COVID-19~\citep{muratov2021critical}. The search for more effective drugs is becoming more urgent but remains underexplored. De Novo Drug Discovery incurs billions of dollars in costs and still confronts a high failure rate in its early stages~\citep{tong2021generative}. Drug Improvement addresses the limitations of De Novo drug discovery by building upon existing FDA-approved drugs. 
The DrugImprover framework~\citep{liu2023drugimprover} leads the way in tackling drug optimization by using Tanimoto similarity~\citep{landrum2013rdkit} to maintain beneficial properties while targeting multiple objectives with its novel Advantage-alignment Policy Optimization (APO) algorithm. It also provides a specialized dataset for optimizing drugs against COVID and cancer proteins.
REINVENT 4~\citep{he2021molecular, he2022transformer, loeffler2024reinvent}, in further, utilize the advanced generative capabilities of Transformers and large language models (LLMs) to address the drug optimization problem.

\begin{figure}[ht]
    \centering
    \begin{subfigure}{0.45\textwidth}
    \centering
    \includegraphics[
    height=1.5cm, 
    clip={0,0,0,0}]{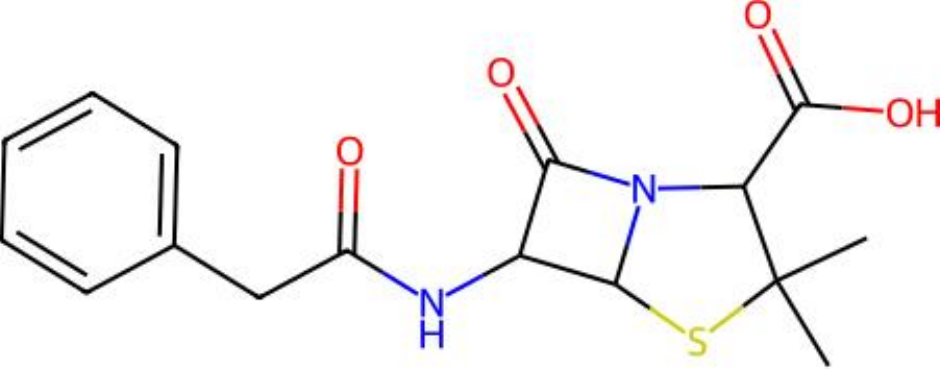}
    \caption{Penicillin}\label{fig:results:vertebral}
\end{subfigure}
\begin{subfigure}{0.45\textwidth}
    \centering
    \includegraphics[
    height=1.5cm, 
    clip={0,0,0,0}]{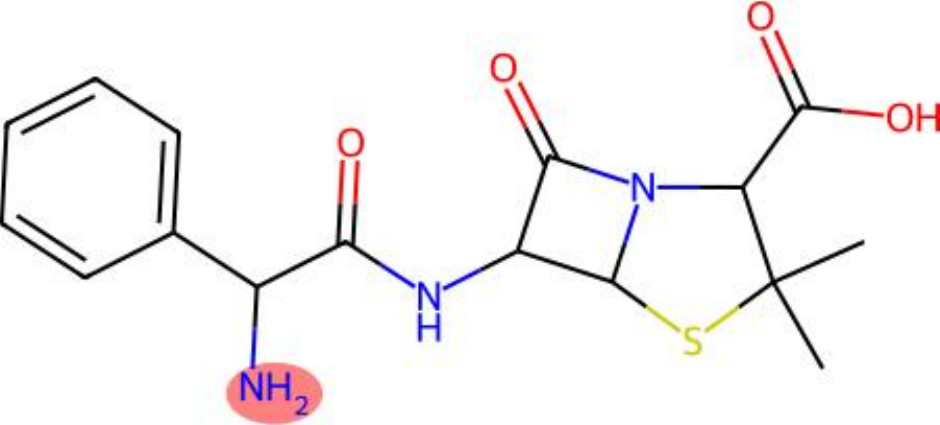}
    \caption{Ampicillin}\label{fig:results:hiv}
\end{subfigure}
    \caption{
    Penicillin in drug optimization. With adding a simple functional group NH2 (in \textcolor{myred}{red}), Ampicillin has resolved the rash side effect bring about by Penicillin.
    }
    \label{fig:penicillin_example}
    \vspace{-0.1cm}
\end{figure}

Although REINVENT 4 has further improved performance, demonstrated promising outcomes, and achieved state-of-the-art performance, its effectiveness is still limited due to the black-box nature of the LLMs' generation process. In this process, LLMs optimize for either maximum likelihood or specific objectives set by users during the decoding phase. More specifically, it cannot specify the generation specific locations or the length of tokens to be generated.
Such limitations are further intensified in drug optimization with REINVENT 4:
(1) If an important substructure exists within the original molecule, REINVENT 4 might miss it and fail to preserve it in the generated ones, even though retaining that substructure could be beneficial.
(2) Real-world examples of drug improvement, such as the addition of an NH2 group to original drugs, demonstrate significant enhancements in effectiveness and reduced side effects. For instance, Ampicillin's modifications over Penicillin, illustrated in Figure \ref{fig:penicillin_example}, show increased activity range, stomach acid resistance, and better absorption while retaining the essential beta-lactam ring crucial for antibiotic activity. Ampicillin's main distinction from Penicillin is its side chain, altered to penetrate gram-negative bacteria's outer membranes more effectively.
(3) If the fragment has associated side effects, drugs derived from it might inherit these issues, contradicting the goal of optimization.
(4) Based on the given drug, we need to decide to what extent (e.g., how many atoms to add) to preserve or alter the original structure.
\vspace{-0.5cm}
\vspace{-0.5cm}

To address the challenges of the GPT model and drug optimization, this work introduces \algname, a Bidirectional Causally Masked Seq2seq GPT tailored for controllable generation in drug optimization. Drawing inspiration from biological processes like growth and evolution, the training of this model leverages the SMILES representation \citep{weininger1988smiles} to enable the expansion, contraction, and mutation of molecule sequences at any point while maintaining essential structures.

This model, trained with a novel causally masked seq2seq objective, 
merges causal, masked, and seq2seq language modeling, enabling comprehensive generative modeling with bidirectional context, facilitating precise modifications without disrupting the molecule’s overall structure.
The model offers controllable capabilities including:
1) Generating new molecules while preserving original beneficial functional groups or scaffolds.
2) Removing atoms or groups causing side effects.
3) Precisely adding new atoms at specified scales and positions, with size hints that guide the generation process.
4) Controlling mutations, allowing expansion or contraction based on sub-sequences to achieve specific molecular configurations.

In summary, our contributions are:

$\bullet$ We propose a Causally Masked Seq2seq (CMS) objective that merges causal, masked, and seq2seq language modeling. CMS enables precise management of specific locations and ranges within a sequence, facilitating expansion, reduction, or mutation over chosen or random lengths, all while preserving the integrity of any specified positions or subsequences.

$\bullet$  We developed \algname completely from the ground up, training it on the CMS objective. This process included designing the training corpus, introducing a novel pre-training strategy, and devising a unique generation process.

$\bullet$  Through extensive experiments and ablation studies on real-world viral and cancer-related benchmarks, we demonstrate the effectiveness and controllability of \algname in outperforming the competing baselines in improving upon existing molecules and drugs across targeted objectives, resulting in superior drug candidates.

\section{Related Work}\label{sec:related}

\subsection{Causally Masked Language Modeling}

\fix{Causal Language Modeling (CLM) is an autoregressive method employed in models such as GPT-4~\citep{achiam2023gpt}, predicting the next token using only prior token information. While effective in applications like text generation~\citep{ li2024pre} and dialogue systems~\citep{hosseini2020simple}, CLM's unidirectional approach is a limitation. 
}
\fix{Masked Language Modeling (MLM), used in models like BERT~\citep{devlin2018bert}, predicts hidden tokens using bidirectional context. Although it processes only about 15\% of tokens during training, limiting some uses, MLM is still broadly applied in biology for representation learning~\citep{chithrananda2020chemberta, lin2023esm}.}
Causally Masked objective improve MLM by providing a type of hybrid of causal and masked language models by enabling full generative modeling while also providing bidirectional context when generating the masked spans. 
However, the existing State-of-the-Art (SOTA) Causally Masked models~\citep{aghajanyan2022cm3} are still limited in enabling the controllable mutation function including conditional expansion and contraction, and the current design of masked tokens leads to misleading interpretations between the masked token and its context. In this work, we address both limitations by redesigning the masked tokens and incorporating a sequence-to-sequence model.

\subsection{Sequence-to-Sequence Modeling }

Seq2Seq, or Sequence-to-Sequence models~\citep{dey2017gate,graves2012long,xue2020mt5,wang2020fairseq, ni2021sentence}, employ an encoder-decoder structure where the encoder interprets the input sequence, and the decoder constructs the output sequence. This method is frequently utilized in tasks such as machine translation~\citep{chen2018best,tiwari2020english,wang2022understanding}, summarization~\citep{prasad2020automatic,shi2021neural}, and question-answering~\citep{tang2018learning,wu2020seq2seq}. Due to their ability to manage complex tasks that require transforming input into output, Seq2Seq models are highly versatile and suitable for a broad spectrum of NLP applications.
Nevertheless, Seq2seq models exhibit limitations in coherence, context understanding, handling variable-length inputs, training efficiency, and capturing bidirectional context when compared to CLMs and MLMs. In this study, we introduce Causally Masked Seq2seq (CMS) modeling, conceptualizing the seq2seq model as a controllable conditional mutation component in biological sequences, and harnessing the strengths of seq2seq models, CLMs, and MLMs.

\subsection{Controllable Generation.}
In the field of computer vision, the introduction of generative adversarial networks~\citep{goodfellow2014generative}  enhanced the quality of image generation. Subsequent research focused on methods to control the generative process and improve the estimation of generative distributions~\citep{kingma2013auto,chen2016infogan,arjovsky2017wasserstein}. In the realm of natural language processing, language models are often developed as conditional models tailored for specific text generation tasks~\citep{brants2007large,sutskever2014sequence,rush2015neural}. Typically, prompts created by models or those written by humans serve merely as a rough starting point for the generated text. This raises questions about how to achieve more explicit control over text generation. Recent advancements in transformer architecture~\citep{vaswani2017attention,radford2019language} and diffusion models~\citep{ho2020denoising} have led to improved control in both text and image generation~\citep{li2019controllable,keskar2019ctrl,raffel2020exploring,li2022diffusion,epstein2023diffusion,zhang2023survey,liang2024controllable}. However, these techniques are not specifically adapted for biological sequences, which may involve unique challenges in nature, such as expansion, reduction, or mutation at specific locations and ranges with desired properties. 
\algname is specifically designed for biological sequences and addresses these challenges by introducing a novel CMS objective.

\subsection{Large Language Models for Drug Optimization}

Large language models have been employed in molecule generation, as evidenced by studies such as MolGPT~\citep{bagal2021molgpt}, C5T5~\citep{rothchild2021c5t5}, and ChemGPT~\citep{frey2023neural}. More recently, ERP~\citep{liu2024erp} has utilized LLMs for drug discovery. In contrast, our work focuses on the drug optimization domain to improve upon existing drugs rather than designing from scratch. In the drug optimization domain, 
DrugImprover~\citep{liu2023drugimprover} starts to effectively define the drug optimization problem by using reinforcement learning with a combination of multiple objectives. Moreover, it integrates Tanimoto similarity~\citep{landrum2016rdkit} as an additional term in the rewards function to ensure that the RL-fine-tuned model generates molecules similar to existing drugs. However, DrugImprover employs an LSTM as the generative model, which has limitations in scalability, capacity, and contextual understanding.
Reinvent 4~\citep{he2021molecular, he2022transformer, loeffler2024reinvent} has made efforts in developing transformer-based generative models and achieving state-of-the-art performance in the Drug Optimization domain with an emphasis on pretraining with simply adopt REINFORCE~\citep{williams1992simple} finetuning. 
Although pretraining aids in producing molecules that resemble those in the training dataset, it naturally limits the scope of exploration because of biases inherent in the training data.
In addition, REINVENT 4 lacks controllability during generation, a crucial aspect for drug optimization. In this study, we tackle controllability issues and surpass the current state-of-the-art, REINVENT 4, in drug optimization benchmarks.

\section{Preliminaries}
\textbf{LLM.} Let $\tokenSeq=\bracket{\token_1, \token_2, \cdots, \token_n}$ be a sequence of tokens representing an input sentence (prompt), where each $\token_i$ is a token from a vocabulary $\Vocabulary$. Let $\mathbf{Y}=\bracket{y_1,y_2,\cdots, y_T}, y_i\in \mathcal{Y}$ be the output sequence of tokens with vocabulary $\mathcal{Y}$.  $\Vocabulary$ and $\mathcal{Y}$ are potentially different vocabularies. Note that $\mathbf{y}_{<t} = \bracket{y_1, \cdots, y_{t-1}}, \mathbf{y}_{T}:=Y$.
$\horizon$ represents the length of sequence. 
Each training corpus begins with a start token $\bracket{\text{BOS}}$, follows with a sequence of tokens $\mathbf{y}$ where each $y_i$ belongs to $\Vocabulary$, and concludes with a termination action $\bracket{\text{EOS}}$.
Each molecule is depicted using a sequence of tokens $\mathbf{y}$ to assemble a SMILES string, applicable to both incomplete and complete molecular structures.
Let us denote $\circ$ as string concatenation, and let $\mathcal{V}^*$ represent the Kleene closure of $\mathcal{V}$.
The set of training corpus $\corpus$ is defined as: 
    $\corpus := \curlybracket{\text{[BOS]} \circ \mathbf{v} \circ \text{[EOS]}~|~\mathbf{v}\in \mathcal{V}^*}.$

The LLM generator policy $\policy_{\theta}$, which is parameterized by a deep neural network (DNN) with learned weights $\theta$, is defined as a product of probability distributions:
$\policy_{\theta}\paren{\mathbf{y}|\mathbf{x}}=\prod_{t=1}^{|\mathbf{y}|} \policy_{\theta}\paren{y_t|\mathbf{x},\mathbf{y}_{<t}}$, where  $\policy_{\theta}\paren{y_t|\mathbf{x},\mathbf{y}_{<t}}=P\paren{y_{t}|\mathbf{y}_{<t},X}$ is a distribution of next token $y_t$.
The text generation decoding process is designed to select the most probable hypothesis from all possible candidates by addressing the following optimization problem:
$\mathbf{y}^{\star}=\argmax_{\mathbf{y}\in \hypothesis_{\horizon}} \log \policy_{\theta}\paren{\mathbf{y}|\mathbf{x}}. $

\textbf{CLM.}
CLM is a variant of language modeling where the model is trained to estimate the probability of $\token_i$ conditioned on the preceding tokens $\mathbf{\tokenSeq}_{<i}$, where $\mathbf{\tokenSeq}_{<i}= \token_1,\token_2,\cdots, \token_{i-1}$, in typically an autoregressively manner.
The objective of CLM is to maximize the log likelihood of observing the correct next token $\token_i$ given all the previous tokens in the sequence $\tokenSeq_{<i}$, which could be formulated as 
$\max_{\theta}\sum^n_{i=1} \log P\paren{x_i|\mathbf{\tokenSeq}_{<i};\theta}$, where $P\paren{x_i|\mathbf{\tokenSeq}_{<i};\theta}$ is the conditional probability of observing token $\token_i$ given all the preceding tokens $\tokenSeq_{<i}$.
Causal Language Modeling is particularly powerful for generating text, as it conditions on all previous tokens, ensuring that each generated word is based on the full history of the text generated so far.

\textbf{MLM.}
In MLM, a subset (around 15\%) of the tokens in $\tokenSeq$ is randomly selected and replaced with a special token [MASK].
Let us denote this masked sequence as $\maskedTokens$ and unmasked sequence as $\unmaskedTokens$, $\unmaskedTokens=\curlybracket{x_i},x_i\in \tokenSeq~\text{and}~ x_i\notin \maskedTokens$.
The objective of the MLM is to predict the original tokens of the masked positions based solely on the unmasked context $\unmaskedTokens$, which can be represented as maximizing the likelihood: $\likelihood_{MLM}=\prod_{i\in \maskedTokens} P\paren{\token_i|\unmaskedTokens;\theta}$, where 
$P\paren{\token_i|\unmaskedTokens;\theta}$ represents the conditional probability of observing token $\token_i$ given the context provided by the unmasked tokens in $\unmaskedTokens$.
$\theta$ represents the parameters of the model.
The parameters  $\theta$ of the model 
are optimized to maximize the likelihood of the correct tokens at the masked positions.
During training, the model learns to utilize the surrounding context to predict the masked tokens, which helps it develop a deep understanding of language structure and usage.
MLM has proven effective for pre-training language models that are later fine-tuned for various downstream tasks.

\textbf{Seq2Seq.}
Sequence-to-sequence (seq2seq) modeling is a framework in natural language processing designed to convert sequences from input sequence to output sequence.
Seq2seq models typically consist of two main components: an encoder and a decoder, with model parameter $\theta_{enc}$ and $\theta_{dec}$ respectively. 
The encoder processes the input sequence $\tokenSeq$ to a fixed-dimensional vector representation $\mathbf{c}$ 
to capture the semantic or contextual information. The decoder's objective is to generate the target sequence $\mathbf{Y}$ given the encoded representation $\mathbf{c}$. 
The objective in training seq2seq models is typically to maximize the log likelihood of the correct output sequence $\mathbf{Y}$ given the input sequence $\tokenSeq$ across a dataset of paired sequences:
$\max_{\theta_{enc},\theta_{dec}}\sum_{\paren{\tokenSeq,\mathbf{Y}}}\log P\paren{\mathbf{Y}|\tokenSeq}$, where $\mathbf{P}$ is product of the conditional probabilities of each output token and
$P\paren{\mathbf{Y}|\tokenSeq}=\prod^n_{j=1}P\paren{y_j|\mathbf{Y}_{<j},\mathbf{c};\theta_{dec}}$.
Training involves adjusting both the encoder and decoder parameters to optimize this objective.
Seq2seq models are powerful because they can handle variable-length input and output sequences and are capable of learning complex transformations between different types of sequence data.

\textbf{Limitation.}
The existing models -CLM, MLM, and seq2seq- have limitations in controllable generation, which is especially important for drug optimization tasks that need to preserve specific structures and allow expansion, shrinking, or mutation at specific positions. The current state of the art in drug optimization, REINVENT 4, although it incorporates various similarity metrics in building the training corpus for pre-training the transformer model, still does not yield ideal results due to a lack of controllability in generation. The beneficial structure of the original drug often fails to preserve. In this work, we propose \algname, which effectively addresses above limitations of current GPT models in controllable drug optimization.

\section{\algname}

In this section, we propose \algname, a ground-up designed GPT model for molecule optimization. We first introduce the novel Causally Masked Seq2seq (CMS) Objective as the foundation of \algname. Then, we discuss the design of GPT, including designing the training corpus, a pre-training strategy, and a generation process.

\subsection{Causally Masked Seq2seq (CMS) Objective.}\label{sec:cm_obj}

\begin{figure}[!ht]
    \centering
    \includegraphics[width=1\linewidth]{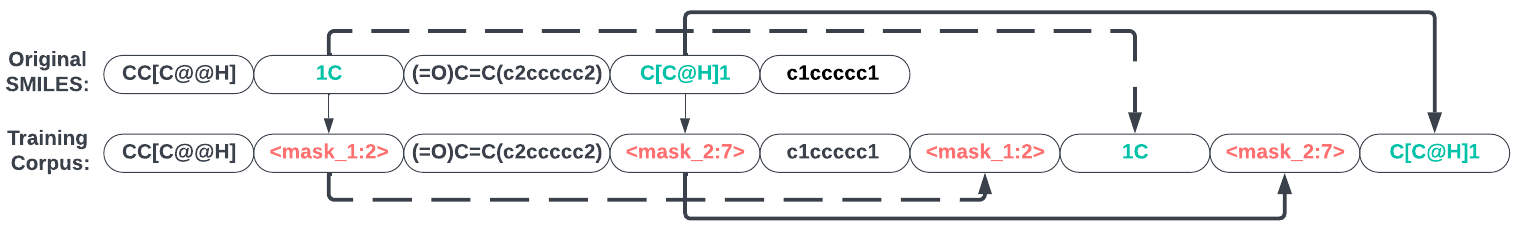}
    \caption[A visual representation of causal masked objective on molecule incorporated with size hints]
    {The visual representation of our causal masked objective on a molecule features two mask spans $(n=2)$, each with a specific size hint. The first span, <$mask\_1:2$>, covers two tokens, and the second, <$mask\_2:7$>, covers seven tokens.}
    \label{fig:cm_mol_example_with_sizeHint}
\end{figure}

Masked, causal, and seq2seq language modeling each offer unique benefits and limitations. Masked models encode bi-directional contexts but only decode about 15\% of the tokens during training. Causal models, being decoder-only, process every token but are restricted to left-to-right contexts. Seq2seq models are versatile yet often lack bidirectional context and precise generation control. To combine the strengths of MLM, CLM, and seq2seq models and draw inspiration from biological molecule evolution using SMILES representation—which allows for molecular expansion, shrinking, and mutation—we introduce the Causally Masked Seq2seq (CMS) Objective.
The CMS objective enables per-token generation, incorporating optional bidirectional and seq2seq functionality for greater adaptability. It allows for precise control over specific positions and spans within sequences, supporting the expansion, contraction, or mutation of segments while maintaining the integrity of designated areas. The construction of the CMS objective involves the following steps:

\paragraph{Designing the corpus.} Our methodology for developing the CMS objective to a SMILES~\citep{weininger1988smiles} string of length $\len$ begins with the most basic corpus suitable for the CLM objective as
$\corpus=\curlybracket{[BOS],{x_1,\cdots,x_{\horizon}},[EOS]}$.
\paragraph{Blending the MLM objective.} We then
build the MLM objective on top of CLM. It involves a probability $\prob$ to determine the total number of tokens to mask as $\lfloor{\len\cdot \prob}\rfloor$. Let us denote $N\in \mathbb{R}^{+}$ as the number of span of mask in the source document. 
\begin{equation}
{[BOS],x_1,\cdots,
\underbrace{x_{idx_1},\cdots, x_{idx_1+\lfloor \len \cdot \prob \rfloor}}_{<mask\_n:\len\cdot\prob>},\cdots,
x_{\horizon},[EOS]},
\end{equation}
For $N=1$, let us choose a random starting index $idx_1 \sim \bracket{0, \len - \lfloor{\len\cdot \prob}\rfloor - 1}$, and proceed to mask tokens in range $\bracket{idx_1,idx_1 + \lfloor{\len\cdot \prob}\rfloor}$. 
For $N=2$, we divide $\lfloor{\len\cdot \prob}\rfloor$ into two segments, $m1$ and  $m2$, ensuring $m1 + m2 = \lfloor{\len\cdot \prob}\rfloor$ and that each segment's length is uniformly selected from the range $[1,\lfloor{\len\cdot \prob}\rfloor]$.
We then identify a starting point $idx_1$ within $\bracket{0, \len  - \lfloor{\len\cdot \prob}\rfloor - 1}$ for the first mask span and a second starting point $idx_2$ from the range $\bracket{idx_1 + m1 + 1, \len - m2 - 1}$ for the second mask span, ensuring that the two masked segments are non-overlapping and sequentially ordered in the SMILES string. Following the same strategy for selection and masking of these segments, we could reach for any $N$. For the $n_{th}$ span of mask, we replace the span by the token <$mask\_n: \len\cdot\prob$>, where $n$ and $\len\cdot\prob$ represents for the $n_{th}$ masked segment with size hint length $\len\cdot\prob$, which specify the desired length of text to generate for replacing the mask conditioning on tokens length.
Finally, we  reposition the masked spans to the end of the SMILES string, maintaining their sequence order as illustrated in \figref{fig:cm_mol_example_with_sizeHint}. 
In this work, we embed the size hint within the mask token as <$mask\_i:n$> to avoid the ambiguity seen in prior works~\citeauthor{aghajanyan2021size_hint}  that use <$mask\_i$>$n$. This format prevents misinterpretation by models, as numerical values in chemical structures can indicate ring closures or chain lengths.

\paragraph{Blending the seq2seq objective.} 
Finally, we establish CMS objective by applying seq2seq objective on top of MLM and CLM. Initially, we train a GPT model using the MLM and CLM objectives, denoted as $\policy_{\text{CM}}$.
Given a SMILES string, we randomly mask a seq2seq span starting at position $s_1$ and of length $\len$, ensuring it does not overlap with previously masked spans, while regard the remaining tokens as $\mathbf{Z}$. Our goal is to transform this s2s span $\left[x_{s_1}, \cdots, x_{s_1+\len}\right]$ into a target span with desired length $\len^t$. To create this training corpus, we utilize $\policy_{\text{CM}}$ to generate $\len^t$ tokens $\left[m_{1}, \cdots, m_{\len^t}\right]$ with regarded to $\mathbf{Z}$. We then construct the training corpus by mapping the s2s span to the subsequence generated by $\policy_{\text{MLM}}$.
\begin{align}
&x_1,\cdots,x_{s_1-1}, \tuple{\text{mask}\_1:\len^t}, 
\underbrace{x_{s_1+ \len +1 },\cdots, {\color{black}{x_{\horizon}, \tuple{\text{mask}\_1:\len^t}}}}_{\text{Prompt based on the pretrained model in previous step}}
\rightarrow[m_1,\cdots,m_{\len^t}] \nonumber
\end{align}
\begin{align}
&x_1,\cdots,{\tuple{\text{s2s}\_i\_\len^t:x_{s_1},\cdots,x_{s_1+\len}}},
\underbrace{\cdots, {\color{black}x_{\horizon}, {\tuple{\text{s2s}\_i\_\len^t:x_{s_1},\cdots,x_{s_1+\len}}},[m_1,\cdots,m_{\len^t}]}}_{\text{Training corpus for seq2seq objective}} \nonumber
\end{align}
where ${\tuple{\text{s2s}\_i\_\len^t:x_{s_1},\cdots,x_{s_1+\len}}}$ denotes the seq2seq objective conditioned on a specific subsequence $x_{s_1},\cdots,x_{s_1+\len}$ and its bidirectional unmasked tokens. 
The index $i$ indicates the $i$-th span, and $\len^t$ represents the target length of the generated subsequence. 
Unlike conventional sequence-to-sequence models, our work on seq2seq is also conditioned on and benefits from the bidirectional context surrounding the seq2seq span.
This approach allows for the incorporation of task-specific length priors into prompts, resulting in outputs that are more precise and controlled.

\begin{figure}[!ht]
    \centering
    \includegraphics[width=1\linewidth]{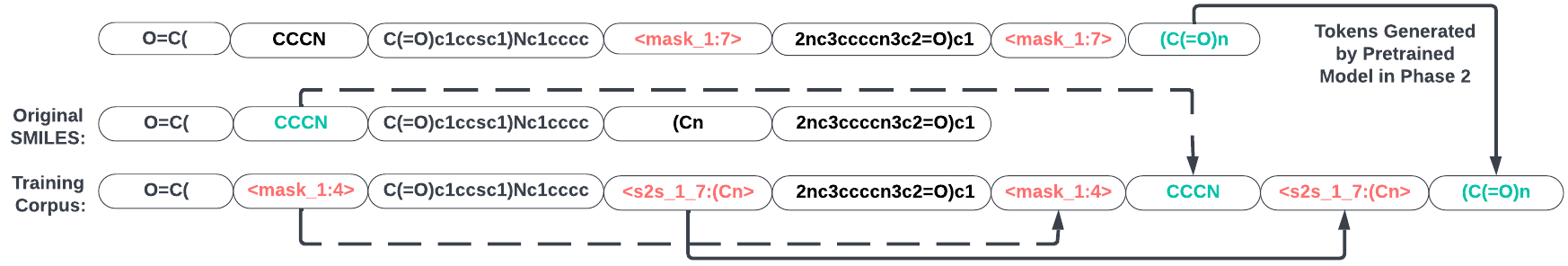}
    \caption
    {The visual representation of building the training corpus with both masked and seq2seq spans for seq2seq causal masked objective. }
    \label{fig:s2s_cm}
\end{figure}

\subsection{The Design of the Controllable GPT.}

\paragraph{Pretraining.}
In this work, 
we propose a novel three-phase training approach to train a GPT model under CMS objective.

\textit{In the initial phase.} Our objective is to train a LLM specifically designed for understanding molecules. This training employs a CLM approach. CLM is an autoregressive technique where the model learns to predict the next token in a sequence based solely on the preceding tokens. This creates a unidirectional context model, which means it only considers past information and ignores any future context when making predictions. For this phase, the model is trained on a dataset comprised of texts about ligands. This dataset enables the model to accurately learn the representation of compounds, including their chemical structures and properties.

\textit{In the second phase.} 
Building on the success of the LLM developed in Phase 1, which demonstrated high accuracy in generating molecular structures, we proceed to refine the model’s training. This phase employs a causally masked objective with multiple mask tokens, each with a size hint, as illustrated in Figure \ref{fig:cm_mol_example_with_sizeHint}.
In this phase, the model, denoted as $\policy_{CM}$, benefits from both Causal Language Modeling (CLM) and Masked Language Modeling (MLM), which enhance CLM's performance by utilizing bidirectional context. $\policy_{CM}$ is capable of generating molecules in a controlled manner, specifying both the target length and the position for expansion.

\textit{In the third phase.} 
Ultimately, we achieve building the GPT under CMS objective by further refining the causally masked model,$\policy_{CM}$, through the integration of a sequence-to-sequence objective. We trained our model, denoted as $\policy_{CMS}$, using the training corpus outlined in \figref{fig:s2s_cm} to refine the causally masked model $\policy_{CM}$ developed in Phase 2.
This advancement aims to enhance the model's controllable generation in terms of both contraction and mutation. It mimics the mutation behavior in biological sequences. Thus, our $\policy_{SCM}$ achieves controllable generation in expansion, contraction, and mutation at specific positions or ranges, in either a random or specified length.

\textit{Loss function.} 
Instead of altering the standard cross-entropy loss \XL{detail}to consider the loss from predicting masked tokens negligible, we treat masked tokens like regular tokens, subject to the usual loss calculations. This method is used because our training data may contain multiple masked tokens, each with size hint information indicating the number of tokens to generate in place of the mask. Thus, it's crucial to accurately predict both the presence of these masked tokens and their corresponding size hints.

\begin{figure}[!ht]
        \centering
    \includegraphics[width=1\linewidth]{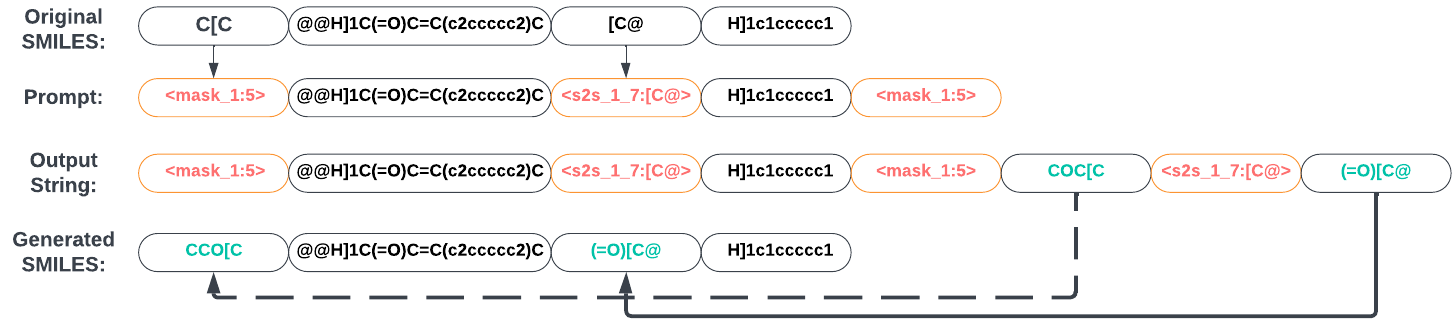}
        \caption{Modification of an original molecule. This figure illustrates the process of altering a molecule's structure. Key steps include replacing original segments with masked and sequence-to-sequence tokens (highlighted in \textcolor{myred}{red}), generating new molecular segments (in \textcolor{ForestGreen}{green}) by the model, and manually reintegrating these segments into the molecule. }
        \label{fig:drugimprovercm_gen_ex1}
    \end{figure}

\begin{figure}[!ht]
        \centering
        \includegraphics[width=1\linewidth]{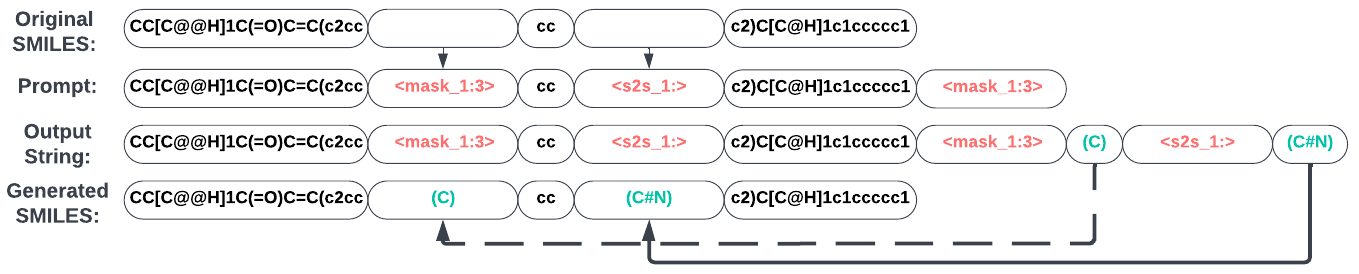}
        \caption{Expansion of an original molecule: Mask tokens (in \textcolor{myred}{red}) are inserted into the SMILES string, prompting the generation of new segments (in \textcolor{ForestGreen}{green}). These segments are then manually added to the molecule, showcasing the model's capability to expand molecular structures both creatively and precisely.}
        \label{fig:drugimprovercm_gen_ex2}
    \end{figure}

\paragraph{Generation Process.} The prompt, output string, and generated SMILES for \algname can be viewed in figure \ref{fig:drugimprovercm_gen_ex1} and figure \ref{fig:drugimprovercm_gen_ex2}. More specifically,
in the process of generating new molecular structures, \algname employs a method that either modifies existing molecules or adds new elements to them without altering the original essential structure, showcasing the flexibility and precision of the model in generating novel molecular designs. This is illustrated through two examples:

\textit{Modifying the Original Molecule:} Initially, two segments of the original molecule's SMILES string are identified and replaced with mask and seq2seq token respectively, which are placeholders indicating where and how long the new segments should be. 
These mask tokens are then processed by the model, which generates new segments in their place. The generated segments, highlighted in green, are manually repositioned to replace the original masked segments, effectively changing the molecule's structure and construct a new molecule. 
This process is depicted in \figref{fig:drugimprovercm_gen_ex1}, where the mask and seq2seq token are shown in red and the newly generated segments in green. 
The caption for \figref{fig:drugimprovercm_gen_ex1} explains this process in detail, emphasizing the manual reintegration of generated tokens.

\textit{Adding to the Original Molecule Without Modification:} In this scenario, instead of replacing parts of the SMILES string, one mask token and one seq2seq token are inserted at random positions within the string. These tokens serve as prompts for the model to generate new molecular segments that are then manually inserted into the specified positions, expanding the original molecule without altering its existing structure. This approach is visualized in \figref{fig:drugimprovercm_gen_ex2}, with the mask tokens again represented in green and the generated segments in red. The caption for \figref{fig:drugimprovercm_gen_ex2} provides a clear explanation of this additive process.

\vspace{-0.2cm}

\section{{Experiments}}\label{experiments}

\paragraph{The language model.} 
We employ the Byte Pair Encoding (BPE) method \citep{gage1994new_bpe, sennrich2015neural_bpe} to initially pre-train our tokenizer using raw SMILES strings, and GPT-2-like Transformers for causal language modeling. We use the standard 11M Drug-like Zinc dataset for training, excluding entries with empty scaffold SMILES. The dataset is divided into a 90/10 split for training and validation, respectively. (see~\appref{app:training_detail} for more details).
\paragraph{Dataset.}
We employ, from the most recent Cancer and COVID dataset of \citet{liu2023drugimprover}, 1 million compounds from the ZINC15 dataset docked to the 3CLPro~(PDB ID: 7BQY) protein associated with SARS-CoV-2 and the RTCB (PDB ID: 4DWQ) human cancer protein.

\begin{table*}[t!]
\setlength{\tabcolsep}{4pt}
   \centering
    {\small
    \scalebox{0.6}{
    \begin{tabular}{l l c c c c c c c c c c }
        \toprule
        \textbf{Target} %
        & \textbf{Algorithm}
        & {\makecell[c]{Avg \\ Norm Reward~$\uparrow$}}
        & {\makecell[c]{Avg Top 10 \% \\ Norm Reward~$\uparrow$}}
        & {\makecell[c]{Docking ~$\downarrow$}}
        & {\makecell[c]{Druglikeliness ~$\uparrow$}}
        & {\makecell[c]{Synthesizability ~$\downarrow$}}
        & {\makecell[c]{Solubility ~$\uparrow$}}
        & {\makecell[c]{Similarity~$\uparrow$}}
        
        \\
        \midrule
        \makecell[l]{\textbf{3CLPro}} %
        &  \textbf{\makecell[l]{Original}}
        &  \makecell[l]{0.532}
        &  \makecell[l]{{0.689}}
        &  \makecell[l]{-8.698}
        &  \makecell[l]{0.682}
        &  \makecell[l]{3.920}
        &  \makecell[l]{2.471}
        &  \makecell[l]{-}
        \\
        (PDBID:
        &  \textbf{\makecell[l]{MMP \citep{loeffler2024reinvent}}}
        &  \makecell[l]{0.628 $\pm$ 0.001}
        &  \makecell[l]{0.718 $\pm$ 0.000}
        &  \makecell[l]{-8.259 $\pm$ 0.004}
        &  \makecell[l]{0.691 $\pm$ 0.001}
        &  \makecell[l]{2.682 $\pm$ 0.004}
        &  \makecell[l]{3.109 $\pm$ 0.020}
        &  \makecell[l]{0.862 $\pm$ 0.000}
        \\
       \ 7BQY)
        &  \textbf{\makecell[l]{Similarity ($\geq$ 0.5) \citep{loeffler2024reinvent}}}
        &  \makecell[l]{0.615 $\pm$ 0.000}
        &  \makecell[l]{0.706 $\pm$ 0.001}
        &  \makecell[l]{-8.165 $\pm$ 0.024}
        &  \makecell[l]{0.697 $\pm$ 0.004}
        &  \makecell[l]{2.621 $\pm$ 0.006}
        &  \makecell[l]{3.180 $\pm$ 0.029}
        &  \makecell[l]{0.782 $\pm$ 0.001}
        \\
        \textbf{ }
        &  \textbf{\makecell[l]{Similarity ([0.5, 0.7)]) \citep{loeffler2024reinvent}}}
        &  \makecell[l]{0.612 $\pm$ 0.001}
        &  \makecell[l]{0.701 $\pm$ 0.001}
        &  \makecell[l]{-8.187 $\pm$ 0.010}
        &  \makecell[l]{0.691 $\pm$ 0.001}
        &  \makecell[l]{\underline{2.611} $\pm$ 0.009}
        &  \makecell[l]{3.240 $\pm$ 0.014}
        &  \makecell[l]{0.756 $\pm$ 0.003}
        \\
        \textbf{ }
        &  \textbf{\makecell[l]{Similarity ($\geq$ 0.7) \citep{loeffler2024reinvent}}}
        &  \makecell[l]{0.628 $\pm$ 0.001}
        &  \makecell[l]{0.718 $\pm$ 0.001}
        &  \makecell[l]{-8.214 $\pm$ 0.002}
        &  \makecell[l]{0.691 $\pm$ 0.002}
        &  \makecell[l]{2.717 $\pm$ 0.002}
        &  \makecell[l]{3.080 $\pm$ 0.016}
        &  \makecell[l]{0.881 $\pm$ 0.002}
        \\
        \textbf{ }
        &  \textbf{\makecell[l]{Scaffold \citep{loeffler2024reinvent}}}
        &  \makecell[l]{0.602 $\pm$ 0.001}
        &  \makecell[l]{0.703 $\pm$ 0.002}
        &  \makecell[l]{-8.116 $\pm$ 0.002}
        &  \makecell[l]{0.695 $\pm$ 0.001}
        &  \makecell[l]{2.728 $\pm$ 0.008}
        &  \makecell[l]{2.968 $\pm$ 0.038}
        &  \makecell[l]{0.776 $\pm$ 0.001}
        \\
        \textbf{ }
        &  \textbf{\makecell[l]{Scaffold Generic \citep{loeffler2024reinvent}}}
        &  \makecell[l]{0.617 $\pm$ 0.001}
        &  \makecell[l]{0.710 $\pm$ 0.002}
        &  \makecell[l]{-8.179 $\pm$ 0.012}
        &  \makecell[l]{0.701 $\pm$ 0.000}
        &  \makecell[l]{2.645 $\pm$ 0.008}
        &  \makecell[l]{3.090 $\pm$ 0.029}
        &  \makecell[l]{0.801 $\pm$ 0.000}
        \\
        \textbf{ }
        &  \textbf{\makecell[l]{DrugImprover \citep{liu2023drugimprover}}}
        &  \makecell[l]{0.432 $\pm$ 0.002} 
        &  \makecell[l]{0.493 $\pm$ 0.005} 
        &  \makecell[l]{-6.726 $\pm$ 0.007}
        &  \makecell[l]{0.506 $\pm$ 0.002}
        &  \makecell[l]{\textbf{1.306} $\pm$ 0.010}
        &  \makecell[l]{2.057 $\pm$ 0.011}
        &  \makecell[l]{0.531 $\pm$ 0.002}
        \\
        \textbf{ }
        &  \textbf{\makecell[l]{Molsearch \citep{sun2022molsearch}}}
        &  \makecell[l]{0.616 $\pm$ 0.001} 
        &  \makecell[l]{0.726 $\pm$ 0.002} 
        &  \makecell[l]{-8.855 $\pm$ 0.040}
        &  \makecell[l]{0.686 $\pm$ 0.001}
        &  \makecell[l]{3.105 $\pm$ 0.006}
        &  \makecell[l]{2.452 $\pm$ 0.008}
        &  \makecell[l]{\textbf{0.969} $\pm$ 0.001}
        \\
        \textbf{ }
        &  \textbf{\makecell[l]{MIMOSA \citep{fu2021mimosa}}}
        &  \makecell[l]{0.622 $\pm$ 0.001} 
        &  \makecell[l]{\underline{0.734} $\pm$ 0.002} 
        &  \makecell[l]{-8.800 $\pm$ 0.015}
        &  \makecell[l]{0.677 $\pm$ 0.004}
        &  \makecell[l]{3.105 $\pm$ 0.008}
        &  \makecell[l]{2.711 $\pm$ 0.010}
        &  \makecell[l]{\underline{0.959} $\pm$ 0.001}
        \\
        \textbf{ }
        &  \fix{ \textbf{\makecell[l]{DrugEx v3 \citep{liu2023drugex}}}}
        &  \makecell[l]{0.524 $\pm$ 0.001} 
        &  \makecell[l]{0.613 $\pm$ 0.001} 
        &  \makecell[l]{-8.089 $\pm$ 0.013}
        &  \makecell[l]{0.583 $\pm$ 0.002}
        &  \makecell[l]{3.095 $\pm$ 0.005}
        &  \makecell[l]{3.932 $\pm$ 0.008}
        &  \makecell[l]{{0.495} $\pm$ 0.001}
        \\
        \textbf{ }
        &  \textbf{\makecell[l]{\algname (masks only)}}
        &  \makecell[l]{\underline{0.668} $\pm$ 0.001} 
        &  \makecell[l]{\textbf{0.743} $\pm$ 0.001} 
        &  \makecell[l]{\underline{-9.083} $\pm$ 0.003}
        &  \makecell[l]{\textbf{0.718} $\pm$ 0.001}
        &  \makecell[l]{2.750 $\pm$ 0.001}
        &  \makecell[l]{\underline{3.630} $\pm$ 0.005}
        &  \makecell[l]{0.889 $\pm$ 0.001}
        
        \\
        \textbf{ }
        &  \textbf{\makecell[l]{\algname (mask + s2s)}}
        &  \makecell[l]{\textbf{0.671} $\pm$ 0.001} 
        &  \makecell[l]{\textbf{0.743} $\pm$ 0.001} 
        &  \makecell[l]{\textbf{-9.150} $\pm$ 0.001}
        &  \makecell[l]{\underline{0.714} $\pm$ 0.001}
        &  \makecell[l]{2.763 $\pm$ 0.002}
        &  \makecell[l]{\textbf{3.672} $\pm$ 0.003}
        &  \makecell[l]{{0.895} $\pm$ 0.001}
        \\
        \bottomrule
        \textbf{RTCB}
        &  \textbf{\makecell[l]{Original}}
        &  \makecell[l]{0.536}
        &  \makecell[l]{{0.698}}
        &  \makecell[l]{-8.572}
        &  \makecell[l]{0.709}
        &  \makecell[l]{3.005}
        &  \makecell[l]{2.299}
        &  \makecell[l]{-}
        \\
        (PDBID:
        &  \textbf{\makecell[l]{MMP \citep{loeffler2024reinvent}}}
        &  \makecell[l]{0.636 $\pm$ 0.000}
        &  \makecell[l]{0.731 $\pm$ 0.001}
        &  \makecell[l]{-8.465 $\pm$ 0.021}
        &  \makecell[l]{0.709 $\pm$ 0.001}
        &  \makecell[l]{2.599 $\pm$ 0.004}
        &  \makecell[l]{3.013 $\pm$ 0.013}
        &  \makecell[l]{0.845 $\pm$ 0.001}
        \\
        \ 4DWQ)
        &  \textbf{\makecell[l]{Similarity ($\geq$ 0.5) \citep{loeffler2024reinvent}}}
        &  \makecell[l]{0.626 $\pm$ 0.000}
        &  \makecell[l]{0.723 $\pm$ 0.001}
        &  \makecell[l]{-8.511 $\pm$ 0.012}
        &  \makecell[l]{0.713 $\pm$ 0.002}
        &  \makecell[l]{2.543 $\pm$ 0.002}
        &  \makecell[l]{3.082 $\pm$ 0.031}
        &  \makecell[l]{0.760 $\pm$ 0.000}
        \\
        \textbf{ }
        &  \textbf{\makecell[l]{Similarity ([0.5, 0.7)]) \citep{loeffler2024reinvent}}}
        &  \makecell[l]{0.622 $\pm$ 0.001}
        &  \makecell[l]{0.718 $\pm$ 0.000}
        &  \makecell[l]{-8.486 $\pm$ 0.021}
        &  \makecell[l]{0.713 $\pm$ 0.003}
        &  \makecell[l]{2.542 $\pm$ 0.005}
        &  \makecell[l]{3.101 $\pm$ 0.005}
        &  \makecell[l]{0.740 $\pm$ 0.001}
        \\
        \textbf{ }
        &  \textbf{\makecell[l]{Similarity ($\geq$ 0.7) \citep{loeffler2024reinvent}}}
        &  \makecell[l]{0.639 $\pm$ 0.000}
        &  \makecell[l]{0.734 $\pm$ 0.001}
        &  \makecell[l]{-8.496 $\pm$ 0.009}
        &  \makecell[l]{ 0.718 $\pm$ 0.001}
        &  \makecell[l]{2.628 $\pm$ 0.001}
        &  \makecell[l]{2.868 $\pm$ 0.003}
        &  \makecell[l]{0.875 $\pm$ 0.002}
        \\
        \textbf{ }
        &  \textbf{\makecell[l]{Scaffold \citep{loeffler2024reinvent}}}
        &  \makecell[l]{0.609 $\pm$ 0.001}
        &  \makecell[l]{0.718 $\pm$ 0.000}
        &  \makecell[l]{-8.508 $\pm$ 0.026}
        &  \makecell[l]{0.711 $\pm$ 0.000}
        &  \makecell[l]{2.627 $\pm$ 0.002}
        &  \makecell[l]{ 2.803 $\pm$ 0.010}
        &  \makecell[l]{0.735 $\pm$ 0.002}
        \\
        \textbf{ }
        &  \textbf{\makecell[l]{Scaffold Generic \citep{loeffler2024reinvent}}}
        &  \makecell[l]{0.625 $\pm$ 0.001}
        &  \makecell[l]{0.722 $\pm$ 0.000}
        &  \makecell[l]{-8.544 $\pm$ 0.009}
        &  \makecell[l]{0.722 $\pm$ 0.002}
        &  \makecell[l]{2.551 $\pm$ 0.010}
        &  \makecell[l]{2.898 $\pm$ 0.005}
        &  \makecell[l]{0.768 $\pm$ 0.004}
        \\
        \textbf{ }
        &  \textbf{\makecell[l]{DrugImprover \citep{liu2023drugimprover}}}
        &  \makecell[l]{0.478 $\pm$ 0.001} 
        &  \makecell[l]{0.618 $\pm$ 0.002} 
        &  \makecell[l]{-8.701 $\pm$ 0.037}
        &  \makecell[l]{0.486 $\pm$ 0.002}
        &  \makecell[l]{\textbf{1.181} $\pm$ 0.010}
        &  \makecell[l]{2.026 $\pm$ 0.013}
        &  \makecell[l]{0.427 $\pm$ 0.001}
        \\
        \textbf{ }
        &  \textbf{\makecell[l]{Molsearch \citep{sun2022molsearch}}}
        &  \makecell[l]{0.625 $\pm$ 0.001} 
        &  \makecell[l]{0.742 $\pm$ 0.001} 
        &  \makecell[l]{-8.747 $\pm$ 0.009}
        &  \makecell[l]{0.719 $\pm$ 0.001}
        &  \makecell[l]{3.012 $\pm$ 0.004}
        &  \makecell[l]{2.273 $\pm$ 0.005}
        &  \makecell[l]{\textbf{0.950} $\pm$ 0.001}
        \\
        \textbf{ }
        &  \textbf{\makecell[l]{MIMOSA \citep{fu2021mimosa}}}
        &  \makecell[l]{0.631 $\pm$ 0.001} 
        &  \makecell[l]{0.749 $\pm$ 0.001} 
        &  \makecell[l]{-8.972 $\pm$ 0.011}
        &  \makecell[l]{0.706 $\pm$ 0.003}
        &  \makecell[l]{3.080 $\pm$ 0.007}
        &  \makecell[l]{2.561 $\pm$ 0.008}
        &  \makecell[l]{\underline{0.945} $\pm$ 0.001}
        \\
        \textbf{ }
        &  \fix{\textbf{\makecell[l]{DrugEx v3\citep{liu2023drugex}}}}
        &  \makecell[l]{0.592 $\pm$ 0.001} 
        &  \makecell[l]{0.668 $\pm$ 0.001} 
        &  \makecell[l]{-8.762 $\pm$ 0.010}
        &  \makecell[l]{0.583 $\pm$ 0.002}
        &  \makecell[l]{\underline{2.488} $\pm$ 0.005}
        &  \makecell[l]{5.827 $\pm$ 0.010}
        &  \makecell[l]{0.393 $\pm$ 0.001}
        \\

        \textbf{ }
        &  \textbf{\makecell[l]{\algname (masks only)}}
        &  \makecell[l]{\underline{0.675} $\pm$ 0.001} 
        &  \makecell[l]{\underline{0.753} $\pm$ 0.001} 
        &  \makecell[l]{\underline{-9.318} $\pm$ 0.002}
        &  \makecell[l]{\textbf{0.752} $\pm$ 0.001}
        &  \makecell[l]{2.674 $\pm$ 0.001}
        &  \makecell[l]{\underline{3.292} $\pm$ 0.002}
        &  \makecell[l]{{0.883} $\pm$ 0.001}
        
        \\
        \textbf{ }
        &  \textbf{\makecell[l]{\algname (mask + s2s)}}
        &  \makecell[l]{\textbf{0.678} $\pm$ 0.001} 
        &  \makecell[l]{\textbf{0.755} $\pm$ 0.001} 
        &  \makecell[l]{\textbf{-9.377} $\pm$ 0.003}
        &  \makecell[l]{\underline{0.751} $\pm$ 0.001}
        &  \makecell[l]{2.688 $\pm$ 0.001}
        &  \makecell[l]{\textbf{3.328} $\pm$ 0.005}
        &  \makecell[l]{{0.890} $\pm$ 0.001}
        \\
        \bottomrule
        \\
    
    \end{tabular}}}
        \caption{
        {\textbf{Main results.} A comparison of eight baselines including Original, six baselines from REINVENT 4 \{MMP, Similarity ($\geq 0.5$), Similarity $\in [0.5,0.7)$, Similarity $\geq 0.7$, Scaffold, Scaffold Generic\}, DrugImprover, Molsearch, MIMOSA, DrugEx v3 and \algname on multiple objectives 
        based on 3CLPro and RTCB datasets. {The top two results are highlighted as \textbf{1st} and \underline{2nd}.  Results are reported for 5 experimental runs.}
        }
        }
        \vspace{-0.5cm}
        \label{exp:main_result}
\end{table*}

\paragraph{Baselines.}
In this study, we use baseline models such as
DrugImprover~\citep{liu2023drugimprover}, which leverages an LSTM-based generator fine-tuned with APO, 
Molsearch \citep{sun2022molsearch}, a search-based strategy utilizing Monte Carlo Tree Search (MCTS) for molecule generation and optimization, MIMOSA \citep{fu2021mimosa}, a graph-based molecular optimization method driven by sampling \fix{and DrugEx v3\citep{liu2023drugex}, which utilizes transformer-based reinforcement learning for scaffold-driven drug optimization}. Additionally, we incorporate the current state of art model, REINVENT 4, proposed by \citet{ he2021molecular, he2022transformer,loeffler2024reinvent}, which trains a transformer to follow the Matched Molecular Pair (MMP)~\citep{kenny2005structure,tyrchan2017matched} guidelines.
Specifically, given a set $\curlybracket{\paren{X,Y,Z}}$, where $X$ represents source molecule,  $Y$ the target molecule, and $Z$ the property change between $X$ and $Y$, the model learns a mapping from $\paren{X, Z} \in \ensuremath{\mathcal{X}} \times \ensuremath{\mathcal{Z}} \implies Y \in \ensuremath{\mathcal{Y}}$ during training. 
REINVENT 4 defined six different kinds of property change $Z$, {including MMP for user-specified changes, different similarity thresholds, and scaffold-based alterations, where molecules share the same scaffold or generic scaffold.} \fix{All baselines are fine-tuned using the cancer and COVID dataset according to their respective fine-tuning methods.}

\begin{table*}[ht!]
\setlength{\tabcolsep}{4pt}
   \centering
    {\small
    \scalebox{0.58}{
    \begin{tabular}{ l l l }
        \toprule
        \textbf{\makecell[l]{Description}}
        &  \makecell[c]{\textbf{Task 1. Adding to the Original Molecule Without Modification}}
        \\
        \textbf{Molecule}
        & {\makecell[l]{\includegraphics[width=0.55\textwidth]{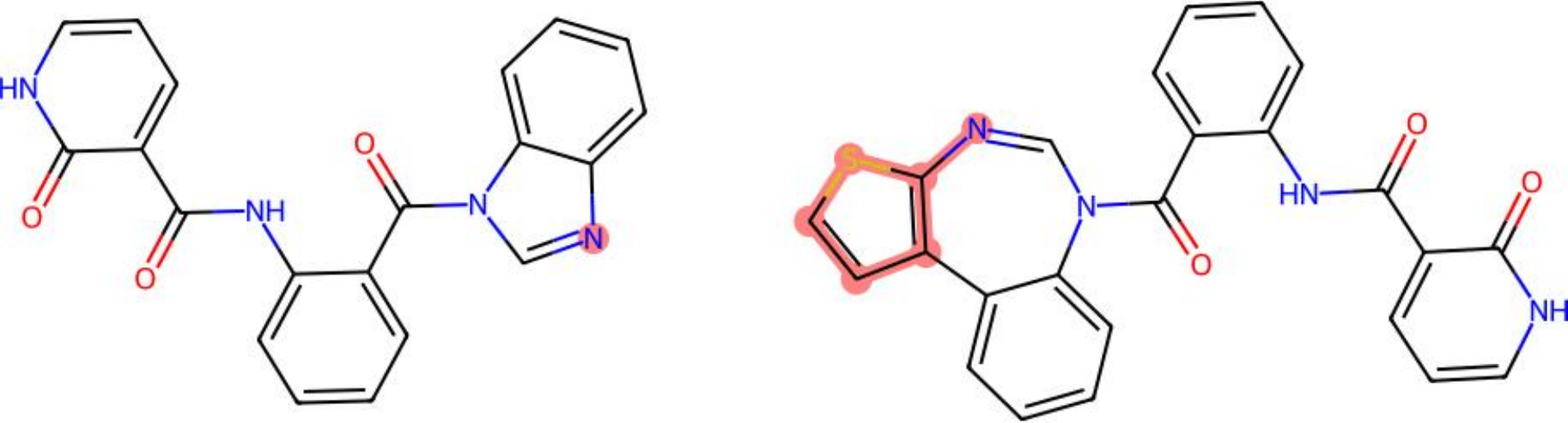}}}
        \\
        \midrule
        \textbf{\makecell[l]{Original SMILES}}
        &  \makecell[l]{O=C(Nc1ccccc1C(=O)n1cnc2ccccc21)c1ccc[nH]c1=O}
        \\
        \midrule
        \textbf{\makecell[l]{Prompt}}
        &  \makecell[l]{O=C(Nc1ccccc1C(=O)n1cnc2{\color{blue}<mask\_1:7>}ccccc21)c1ccc[nH]c1=O{\color{blue}<mask\_1:7>}}
        \\
        \midrule
        \textbf{\makecell[l]{ Masked $\rightarrow$ Generated \\  \text{[token, length]}}}
        &  \makecell[l]{[None, 0] $\rightarrow$ [sccc2c2, 7]}
        \\
        \midrule
        \textbf{\makecell[l]{Generated SMILES}}
        &  \makecell[l]{O=C(Nc1ccccc1C(=O)n1cnc2{\color{green}sccc2c2}ccccc21)c1ccc[nH]c1=O}
        \\
                \toprule
        \textbf{\makecell[l]{Description}}
        &  \makecell[c]{\textbf{Task 2. Modifying the Original Molecule}} 
        \\
        \textbf{Molecule}
        & {\makecell[l]{\includegraphics[width=0.55\textwidth]{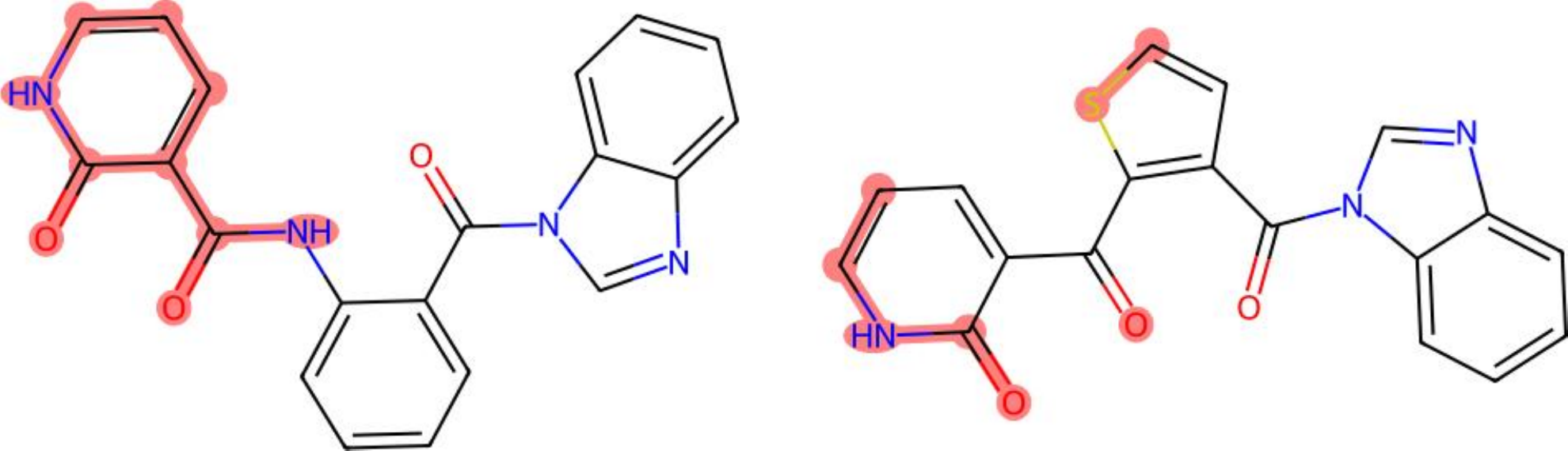}}}
        \\
        \midrule
        \textbf{\makecell[l]{Original SMILES}}
        &  \makecell[l]{O=C(Nc1ccccc1C(=O)n1cnc2ccccc21)c1ccc[nH]c1=O}
        \\
        \midrule
        \textbf{\makecell[l]{Prompt}}
        &  \makecell[l]{O=C({\color{blue}<mask\_1:3>}ccc1C(=O)n1cnc2ccccc21)c1ccc[nH]c1=O{\color{blue}<mask\_1:3>}} 
        \\
        \midrule
        \textbf{\makecell[l]{ Masked $\rightarrow$ Generated \\  \text{[token, length]}}}
        &  \makecell[l]{ [Nc1cc, 5] $\rightarrow$  [c1s, 3] } 
        \\
        \midrule
        \textbf{\makecell[l]{Generated SMILES}}
        &  \makecell[l]{{O=C(\color{green}c1s}ccc1C(=O)n1cnc2ccccc21)c1ccc[nH]c1=O} 
        \\
                \toprule
        \textbf{\makecell[l]{Description}}
        &  \makecell[c]{\textbf{Task 3. Modifying to the Original Molecule: Simplification}}
        \\
        \textbf{Molecule}
        & {\makecell[l]{\includegraphics[width=0.55\textwidth]{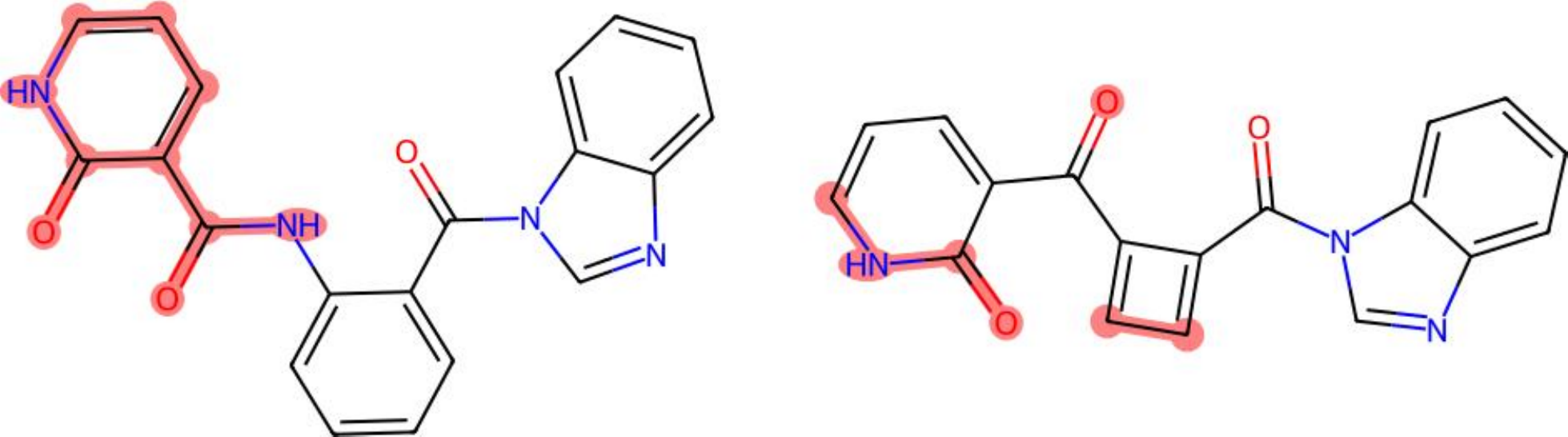}}}
        \\
        \midrule
        \textbf{\makecell[l]{Original SMILES}}
        &  \makecell[l]{O=C(Nc1ccccc1C(=O)n1cnc2ccccc21)c1ccc[nH]c1=O}
        \\
        \midrule
        \textbf{\makecell[l]{Prompt}}
        &  \makecell[l]{O=C({\color{blue}<s2s\_1\_2:Nc1cc>}ccc1C(=O)n1cnc2ccccc21)c1ccc[nH]c1=O{\color{blue}<s2s\_1\_2:Nc1cc>}}
        \\
        \midrule
        \textbf{\makecell[l]{ Masked $\rightarrow$ Generated \\ \text{[token, length]}}}
        &  \makecell[l]{[Nc1cc, 5] $\rightarrow$ [c1, 2]}
        \\
        \midrule
        \textbf{\makecell[l]{Generated SMILES}}
        &  \makecell[l]{O=C({\color{green}c1}ccc1C(=O)n1cnc2ccccc21)c1ccc[nH]c1=O}
        \\
                \toprule
                \textbf{\makecell[l]{Description}}
        &  \makecell[c]{\textbf{Task 4. Modifying the Original Molecule: Expansion} }
        \\
        \textbf{Molecule}
        & {\makecell[l]{\includegraphics[width=0.55\textwidth]{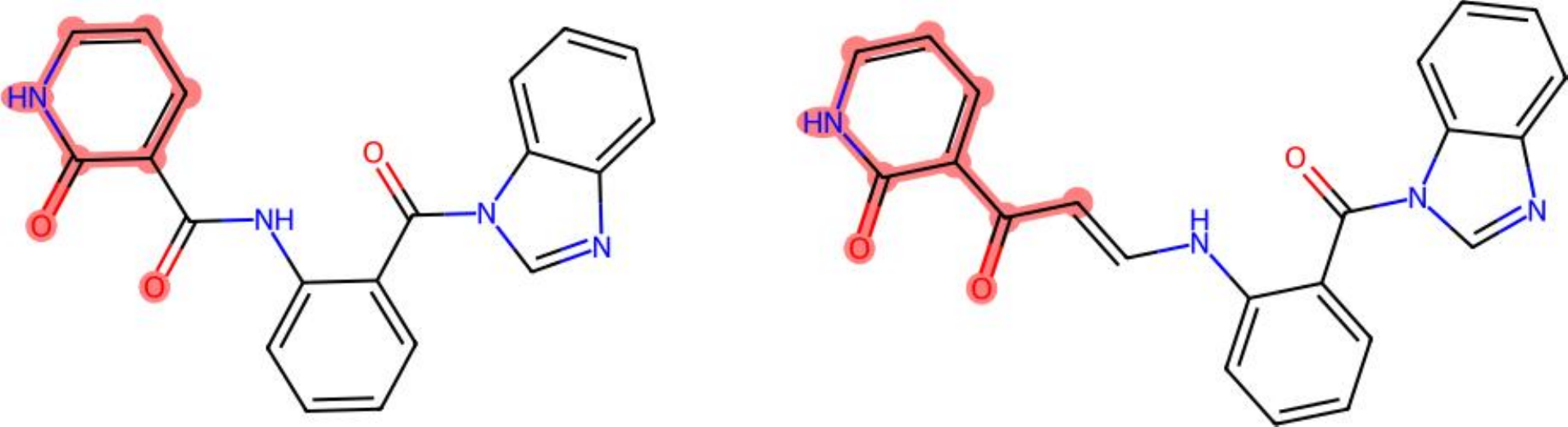}}}
        \\
        \midrule
        \textbf{\makecell[l]{Original SMILES}}
        &  \makecell[l]{O=C(Nc1ccccc1C(=O)n1cnc2ccccc21)c1ccc[nH]c1=O}
        \\
        \midrule
        \textbf{\makecell[l]{Prompt}}
        &  \makecell[l]{O=C({\color{blue}<s2s\_1\_10:Nc1cc>}ccc1C(=O)n1cnc2ccccc21)c1ccc[nH]c1=O{\color{blue}<s2s\_1\_10:Nc1cc>}} 
        \\
        \midrule
        \textbf{\makecell[l]{ Masked $\rightarrow$ Generated \\ \text{[token, length]}}}
        &  \makecell[l]{[Nc1cc, 5] $\rightarrow$ [/C=C/Nc1cc, 10]} 
        \\
        \midrule
        \textbf{\makecell[l]{Generated SMILES}}
        &  \makecell[l]{O=C({\color{green}/C=C/Nc1cc}ccc1C(=O)n1cnc2ccccc21)c1ccc[nH]c1=O} 
        \\
                \toprule
        \textbf{\makecell[l]{Description}}
        &  \makecell[c]{\textbf{Task 5. Modifying to the Original Molecule (Penicillin): Simplification}}
        \\
        \textbf{Molecule}
        & {\makecell[l]{\includegraphics[width=0.55\textwidth]{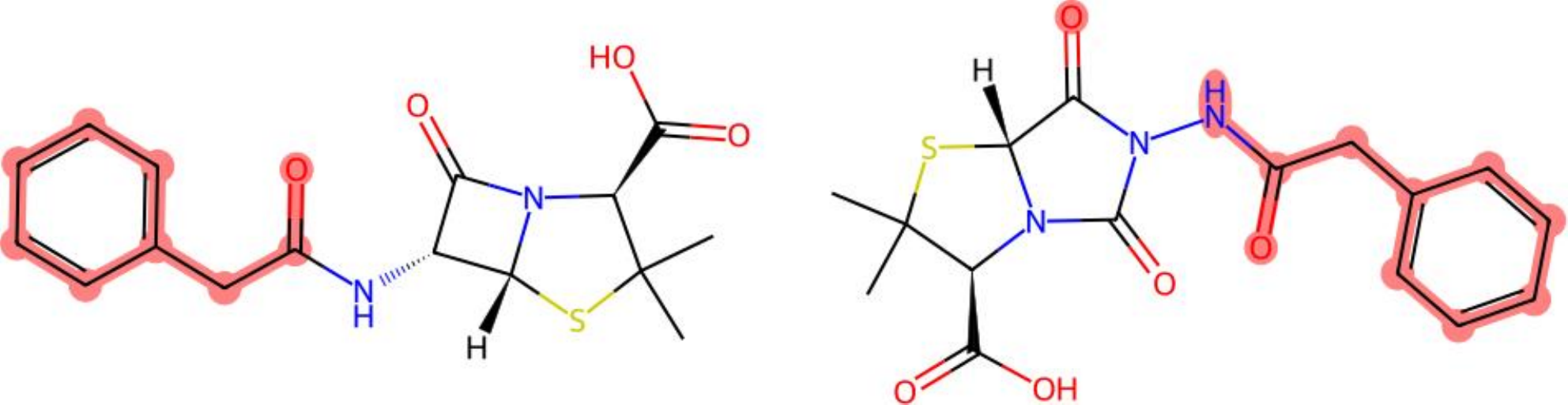}}}
        \\
        \midrule
        \textbf{\makecell[l]{Original SMILES}}
        &  \makecell[l]{CC1([C@@H](N2[C@H](S1)[C@@H](C2=O)NC(=O)CC3=CC=CC=C3)C(=O)O)C}
        \\
        \midrule
        \textbf{\makecell[l]{Prompt}}
        &  \makecell[l]{C1([C@@H](N2[C@H](S1){\color{blue}<mask\_1:6>}(C2=O)NC(=O)CC3=CC=CC=C3)C(=O)O)C{\color{blue}<mask\_1:6>}}
        \\
        \midrule
        \textbf{\makecell[l]{ Masked $\rightarrow$ Generated \\ \text{[token, length]}}}
        &  \makecell[l]{[ [C@@H], 6] $\rightarrow$ [(C2=O), 6] }
        \\
        \midrule
        \textbf{\makecell[l]{Generated SMILES}}
        &  \makecell[l]{CC1([C@@H](N2[C@H](S1){\color{green}C(=O)N}(C2=O)NC(=O)CC3=CC=CC=C3)C(=O)O)C}
        \\
        \midrule
        \textbf{\makecell[l]{Toxicity Score}}
        &  \makecell[l]{ (Original) 2.54 $\rightarrow$ (Generated) 2.35}
        \\
        \bottomrule
                \textbf{\makecell[l]{Description}}
        &  \makecell[c]{\textbf{Task 6.  Modifying the Original Molecule (Penicillin): Expansion}}
        \\
        \textbf{Molecule}
        & {\makecell[l]{\includegraphics[width=0.55\textwidth]{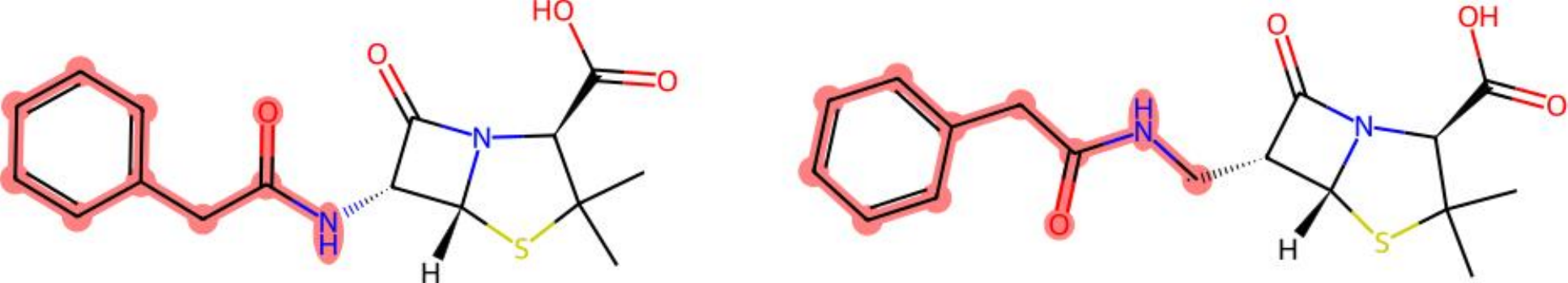}}}
        \\
        \midrule
        \textbf{\makecell[l]{Original SMILES}}
        &  \makecell[l]{CC1([C@@H](N2[C@H](S1)[C@@H](C2=O)NC(=O)CC3=CC=CC=C3)C(=O)O)C}
        \\
        \midrule
        \textbf{\makecell[l]{Prompt}}
        &  \makecell[l]{CC1([C@@H](N2[C@H](S1)[C@@H]{\color{blue}<s2s\_10:(C2=O)>}NC(=O)CC3=CC=CC=C3)C(=O)O)C{\color{blue}<s2s\_10:(C2=O)>}}
        \\
        \midrule
        \textbf{\makecell[l]{ Masked $\rightarrow$ Generated \\ \text{[token, length]}}}
        &  \makecell[l]{ [(C2=O), 6] $\rightarrow$ [(C2=O)[CH], 10]} 
        \\
        \midrule
        \textbf{\makecell[l]{Generated SMILES}}
        &  \makecell[l]{CC1([C@@H](N2[C@H](S1)[C@@H]{\color{green}(C2=O)[CH]}NC(=O)CC3=CC=CC=C3)C(=O)O)C} 
        \\
        \midrule
        \textbf{\makecell[l]{Toxicity Score}}
        &  \makecell[l]{ (Original) 2.54 $\rightarrow$ (Generated) 2.11} 
        \\
        \bottomrule
    \end{tabular}}}
        \caption{\textbf{Ablation studies.} \textbf{Task 1\&2:} Examples using masking and size hints for controllable generation.
        \textbf{Task 3\&4:} Examples using Seq2Seq and size hints for controllable generation.
        \textbf{Task 5\&6:} Examples using \algname to reduce the toxicity of Penicillin while preserving its fundamental structure.
        }
        \label{viz:mask_with_size_hint}
        \vspace{-0.6cm}
\end{table*}

\paragraph{Critics and evaluation metric.} 
\vspace{-0.2cm}
We evaluate seven key attributes for pharmaceutical drug discovery:
1) {\emph{Average normalized reward} is the average of the normalized values of the docking score, drug-likeness, synthesizability, solubility, and similarity across all valid molecules. This is regarded as the most crucial metric.}; 
2) {\emph{Average top 10\% normalized reward} is the average of the normalized reward of the top 10\% of molecules based on their average normalized reward;}
3) \emph{Docking score} (generated, for efficient calculation, with a surrogate docking model: see Appendix \ref{app:surrogate_model})
evaluates the potential of a drug to inhibit the target site.
4) \emph{Druglikeness} assesses the probability of a molecule being a suitable drug candidate;
5) \emph{Synthesizability} measures the synthesizability of a molecule, assigning a score of 1 for easy synthesis and a score of {10} for difficult synthesis {~\citep{ertl2009estimation}}; and
{6) \emph{Similarity} evaluates the similarity between original and generated SMILES using Tanimoto similarity.}

\subsection{Main results}

\tabref{exp:main_result} illustrates the performance comparison between \algname and the competing baseline methods. The results indicate that \algname outperforms the competing baselines across all the metrics except for synthesizability. Notably, \algname achieves the highest Tanimoto similarity score, surpassing both the current state-of-the-art, REINVENT 4, and its six variants. This implies that molecules optimized by \algname not only exhibit structures more similar to the original drug compared to existing methods but also demonstrate improved properties across various metrics. Additionally, when compared to the original baseline, the drugs generated by \algname significantly enhance the original drug across the desired aspects.
These results underscore the superiority and effectiveness of \algname in controllable optimization of original drugs, preserving beneficial structures while optimizing diverse properties. 
In addition, \algname with both masked and seq2seq tokens outperforms the masked token only. This demonstrate that the GPT model developed with our CMS objective surpasses causally masked modeling because of the added capabilities of CMS, such as the controllable mutation function that enables conditional expansion and contraction.

\subsection{Ablation studies}

\vspace{-0.1cm}
\paragraph{Adding to the original molecule without modification.} 
\tabref{viz:mask_with_size_hint} (Task 1) visualizes the addition to the original molecule while preserving the complete original structure. In this experiment, a given original molecule with the SMILES representation \text{O=C(Nc1ccccc1C(=O)n1cnc2ccccc21)c1ccc[nH]c1=O} serves as the basis. Our objective is to extend the ring in the molecule. We designed the prompt by adding a mask token $<mask_1:7>$ to the specific position adjacent to the ring in the SMILES. Finally, we obtained the generated molecule with the desired features (additional ring {\color{red}{in red}}) while maintaining the completeness of the original molecule structure. This study demonstrates the ability of \algname to extend at specific positions with a specific length.

\paragraph{Modifying the Original Molecule.} 
\label{exp:modify}
{In this experiment, our goal is to alter a portion of the original molecule by modifying bonds and atoms connecting the two rings. }
For this purpose, we construct the prompt by substituting the original structure $Nc1cc$ with a masked token $<mask_1:3>$. \tabref{viz:mask_with_size_hint} (Task 2) illustrates the modification of the original molecule by removing the ring and introducing a few atoms, while retaining the majority of the structure. This demonstrate the ability of \algname by modifying partial of molecule and random generated in specific length.

\vspace{-0.1cm}

\paragraph{Conditional Modifying to the Original Molecule: Contraction and Expansion.}
This experiment aims to showcase conditional modifications to the original molecule. Unlike Task 1$\&$2,
where the focus is on modifications and expansion in a random manner, here we concentrate on generating subsequences conditioned on a partial molecule. We undertake two tasks: expanding and shrinking partial molecules based on a given subsequence.
For the simplification task, we successfully reduce a length 5 subsequence, $Nc1cc$, to a length 2 token using the \fix{prompt} token $<s2s\_1\_2:Nc1cc>$. Conversely, for the expansion task, we extend the subsequence to a length of 10 tokens using the  \fix{prompt} token $<s2s\_1\_10:Nc1cc>$. Both tasks yield the desired molecules, as depicted in \tabref{viz:mask_with_size_hint} (Task 3$\&$4).
This demonstrates that \algname is capable of generating molecules controllably for contraction and expansion, conditioned on specific segments of the molecule, to target specific lengths of subsequences.

\fix{
\paragraph{Penicillin Toxicity Reduction.} 

In this study, we utilize the ToxSmi Model~\citep{born2023chemical}, which was trained on the Tox21~\citep{tox21} dataset, encompassing 12 different types of environmental toxicities. The toxicities reported in \tabref{viz:mask_with_size_hint} (Tasks 5$\&$6) represent the sum of 12 toxicity scores. The original molecule, Penicillin, has a predicted toxicity score of 2.54. 
Our proposed controllable methods demonstrate a significant reduction in the toxicity scores of the generated molecule, while preserving the core scaffold structure for preseving desired beneficial properties.

}

\section{Conclusion}\label{sec:con}

In this study, we introduce the novel Causally Masked Seq2Seq (CMS) objective and \algname, which allows precise control over specific sequence areas for expansion, reduction, or mutation while preserving key regions and biologial structure. \algname demonstrated superiority over eight competing baselines in Covid and Cancer drug optimization benchmarks, maintaining high Tanimoto similarity and enhancing drug properties. It also demonstrated its controllability through specific examples in ablation studies. This method highlights \algname's capability for precise generation in drug optimization tasks, despite its limitations. For future directions, we encourage applying \algname in fields beyond our current research scope.

\subsubsection*{Acknowledgements}

This work is supported by the RadBio-AI project (DE-AC02-06CH11357), U.S. Department of Energy Office of Science, Office of Biological and Environment Research, the Improve project under contract (75N91019F00134, 75N91019D00024, 89233218CNA000001, DE-AC02-06-CH11357, DE-AC52-07NA27344, DE-AC05-00OR22725), 
the Exascale Computing Project (17-SC-20-SC), a collaborative effort of the U.S. Department of Energy Office of Science and the National Nuclear Security Administration.

\section*{Impact Statement}
This paper investigates advancements in machine learning for developing controllable language models for molecular optimization. While such models may have certain societal implications, we do not find any that warrant specific emphasis here.

\bibliographystyle{plainnat}
\bibliography{reference}

\clearpage

\appendix

\onecolumn

\section{Appendix}

{\subsection{Pre-training Details}\label{app:training_detail}

We used the ZINC dataset, filtering for Standard, In-Stock, and Drug-Like molecules, resulting in approximately 11 million molecules.

In the second phase of pre-training, we first trained for 10 epochs using a single mask. Subsequently, we trained for another 40 epochs with an equal probability of using either one or two masks. For each epoch, the masks were regenerated to create a more comprehensive masked dataset.

In the third phase of pre-training, we applied different mask configurations with specific probabilities: [one mask (0.1), two masks (0.1), one mask and one seq2seq (0.4), two masks and one seq2seq (0.4)] and train 20 epochs.Similar to the second phase, the masks were regenerated for each epoch to enhance the comprehensiveness of the masked dataset.}

\fix{\subsection{Baselines fine-tuning datasets}

As outlined in Section \ref{experiments}, all baseline models are fine-tuned using the Cancer and COVID dataset, following their respective fine-tuning methodologies. For this process, we utilize one million compounds from the ZINC15 dataset, docked to the 3CLPro protein (PDB ID: 7BQY), which is linked to SARS-CoV-2, and the RTCB protein (PDB ID: 4DWQ), associated with human cancer. These datasets, sourced from the latest Cancer and COVID dataset by \citet{liu2023drugimprover}, are consistently applied across all baselines.

Additionally, these datasets are employed for molecular generation in our proposed methods, with further details on the generation process provided in Section \ref{app:generation}.}

{\subsection{Generation}\label{app:generation}

For each mask and seq2seq, we utilize three random variables: the start index, the number of tokens to be masked, and the number of tokens to be generated. During generation, we apply two settings: [one mask + one seq2seq, and two masks], resulting in a total of six random variables for each setting.

During the generation phase, we randomly sample these six variables 10,000 times, using them as prompts for generation, regardless of whether the generated SMILES are valid or not. In addition, for a given prompt molecule, we adopt TOPPK \citep{liu2024erp} for generation strategy.

After generation, for each prompt molecule/SMILES, we select the top 10 generated molecules/SMILES based on their average normalized reward. The mean of these top 10 molecules/SMILES is then used to obtain the final result for the prompt molecule/SMILES.

 }

{\subsection{Baseline REINVENT 4} \label{app:reinvent}
Following are detailed description of six different kinds of property change $Z$ included in REINVENT 4 \citet{he2022transformer, he2021molecular}
\begin{itemize}[itemsep=0pt,parsep=0pt,topsep=0pt,partopsep=0pt]
    \item \textbf{MMP:} There are user-defined desirable property changes between molecules $X$ and $Y$.
    \item \textbf{Similarity $\geq 0.5$:} The Tanimoto similarity between molecules $X$ and $Y$ is greater than 0.5.
    \item \textbf{Similarity $\in[0.5, 0.7)$:} The Tanimoto similarity between the pair $\left(X,Y\right)$ ranges from 0.5 to 0.7.
    \item \textbf{Similarity $\geq 0.7$:} The Tanimoto similarity between molecules $X$ and $Y$ is greater than 0.7.
    \item \textbf{Scaffold:} Molecules $X$ and $Y$ share the same scaffold.
    \item \textbf{Scaffold generic:} Molecules $X$ and $Y$ share the same generic scaffold.
\end{itemize}}

\subsection{{BPE Tokenization}}\label{app:vocabulary}

Byte Pair Encoding (BPE) is a tokenization algorithm initially designed for data compression and later adapted for use in NLP, particularly in the preprocessing of text for deep learning models. The core idea behind BPE is to iteratively merge the most frequent pair of consecutive bytes (or characters in the context of text) into a single, new byte (or token), thereby reducing the size of the data to be processed. This method has been particularly influential in the development of language models and machine translation systems. 
The BPE method follows these main steps: 
\begin{enumerate}
    \item \textbf{Initial vocabulary preparation: }The text is divided into a sequence of characters or symbols, and a special end-of-word symbol (like <\textbackslash w> or another unique marker) is added to each word to distinguish between the same character sequence occurring within a word and at the end of a word.
    \item \textbf{Frequency Count: } The algorithm counts the frequency of each pair of adjacent characters (or symbols) in the text.
    \item \textbf{Iterative Merging: }
    \begin{itemize}
        \item Identify the most frequent pair of adjacent characters.
        \item Merge this pair into a new single symbol (this does not mean changing the text itself but rather how the algorithm interprets the text).
        \item Update the frequency count of all pairs, considering the newly created symbol.
        \item Repeat this process for a predetermined number of iterations or until a desired vocabulary size is reached.
    \end{itemize}
    \item \textbf{Tokenization: } Once the merging process is complete, the original text can be tokenized (i.e., divided into a sequence of tokens) using the final set of symbols, including the merged ones. This results in a text representation where frequent words or subwords are encoded as single tokens, and less common words are broken down into smaller tokens.
\end{enumerate}

A significant benefit of BPE lies in its capacity to manage rare and out-of-vocabulary words effectively. Since BPE operates at the character level, it can segment words that were not encountered during training, thus reducing the negative effects of unfamiliar words on the model's performance. 
In contexts where tokens of various lengths are randomly masked and relocated to the end of the sequence, as proposed in section \ref{sec:cm_obj}, there's a high likelihood of generating a considerable number of unfamiliar tokens. BPE's approach is particularly beneficial here, as it ensures that the model can still process and understand these novel token sequences by breaking them down into familiar subunits, thereby maintaining robustness and reducing the potential degradation in performance due to unexpected or rare words.

\subsection{Surrogate model}\label{app:surrogate_model}

{The surrogate model~\citep{vasan23} is a simplified version of a BERT-like transformer, widely employed in natural language processing. In this model, tokenized SMILES strings are inputted and then positionally embedded. The outputs are subsequently fed into a series of five transformer blocks, each comprising a multi-head attention layer (with 21 heads), a dropout layer, layer normalization with residual connection, and a feedforward network. The feedforward network consists of two dense layers followed by dropout and layer normalization with residual connection. Following the stack of transformer blocks, a final feedforward network is employed to produce the predicted docking score.} \fix{{ The validation $r^2$ values are 0.842 for 3CLPro dataset and 0.73 for the RTCB dataset.}}

\subsection{Performance scales with mask length}
We conducted an analytical experiment to examine how performance scales with mask length. The results show that the smaller the difference between the length of the generated span and the masked span, the higher the validity will be. In our settings, the validity reaches its highest at 90\% when the length of the generated span is between 5 and 10.

\begin{figure*}[ht!]

    \begin{subfigure}{1\textwidth}
        \centering
        \includegraphics[%
        width=10cm,  clip={0,0,0,0}]{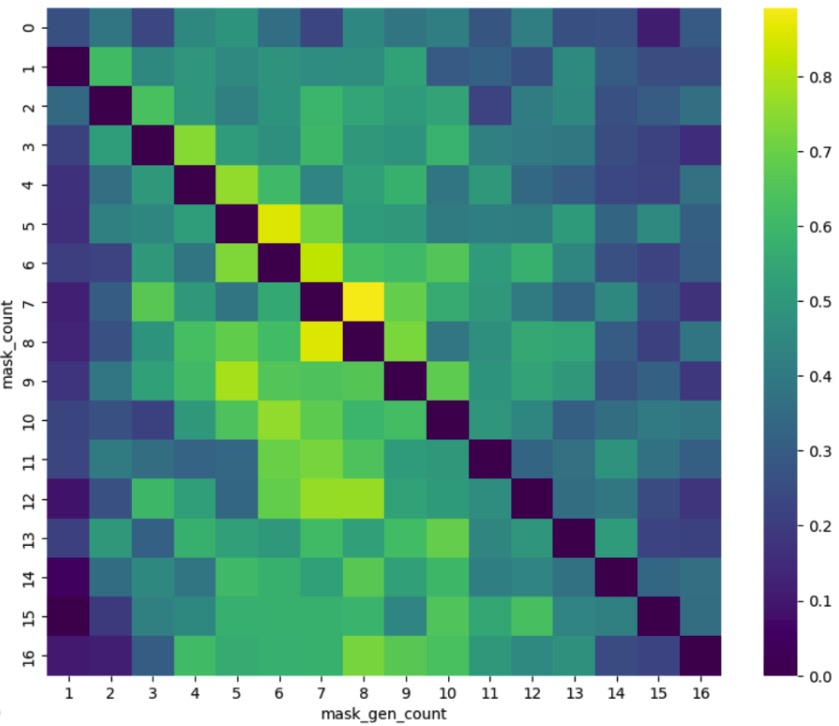}
    \end{subfigure}\hfil
    \caption{The x-axis of this heatmap is the number of tokens that are generated, and the y-axis of this heatmap is the number of tokens that are masked in prompt.  
    }
\end{figure*}

\subsection{{Computing infrastructure {and wall-time comparison}}}\label{app:computing_infrastructure}
{We trained our docking surrogate models using 4 nodes of a supercomputer, each node equipped with CPUs (64 cores) and 4 A100 GPUs. The training time for each model was approximately 3 hours.} 
We performed additional pretraining on a cluster consisting of CPU nodes (approximately 280 cores) and GPU nodes (approximately 110 Nvidia GPUs, ranging from Titan X to A6000, primarily configured in 4- and 8-GPU setups). 

{Pretraining utilizes 8 A100 GPUs, while one single generation uses a single Tesla T4 GPU. Based on the computing infrastructure, {pretraining details as described in Appendix \ref{app:training_detail} and generation details as described in Appendix \ref{app:generation}}, we obtained the wall-time comparison in \tabref{table:wall-time} as follows.}

 \begin{table*}[ht!]
 {
    \centering
    {\scriptsize
    \scalebox{1}{
    \begin{tabular}{l c c  }
        \toprule
        \textbf{}
        & {\makecell[c]{Total Run Time}}
        \\
        \midrule
        \textbf{\makecell[l]{Initial Phase Pretraining}}
        &  \makecell[r]{~18h}
        \\
        \textbf{\makecell[l]{Second Phase Pretraining}}
        &  \makecell[r]{~48h}
        \\
        \textbf{\makecell[l]{Third Phase Pretraining}}
        &  \makecell[r]{~20h}
        \\
        \midrule
        \textbf{\makecell[l]{Generation 10k times \\ for one molecule}}
        &  \makecell[r]{~15mins}
        \\
        \bottomrule
    \end{tabular}}}
    \caption{{Wall-time comparison between different methods.} }
        \label{table:wall-time}   
        }
\end{table*}

\subsection{{Hyperparameters and architectures}}\label{app:hyperparameters}
Table \ref{app:tab:hyperparams} provides a list of hyperparameter settings we used for our experiments.

{For experimentation, 1280 molecules from each of the RTCB and 3CLPro datasets, with docking scores ranging from -14 to -6, are selected. This range is based on \citep{liu2024erp}.}

{In addition, when calculating the average normalized reward for the original molecule, where similarity is not considered, we select the weights for docking, drug-likeness, synthesizability, and solubility as $[0.25] \times 4$.}

\begin{table*}[h!]
    {
    \centering
    {\scriptsize
    \scalebox{1}{
    \begin{tabular}{c c }
        \toprule
        \textbf{Parameter} &  \textbf{Value} 
        \\
        \midrule
        {\makecell[l]{Pretraining}}
        \\
        \midrule
        {\makecell[l]{\quad Learning rate}} &  \makecell[c]{$5 \times e^{-5}$}
        \\
        \midrule
        {\makecell[l]{\quad Batch size}} &  \makecell[c]{$24$}
        \\
        \midrule
        {\makecell[l]{\quad Optimizer}} &  \makecell[c]{Adam}
        \\
        \midrule
        {\makecell[l]{\quad \# of Epochs for Training Initial Phase}} &  \makecell[c]{$10$} \\
        \midrule
        {\makecell[l]{\quad \# of Epochs for Training Second Phase}} &  \makecell[c]{$50$} \\
        \midrule
        {\makecell[l]{\quad \# of Epochs for Training Third Phase}} &  \makecell[c]{$20$}
        \\
        \midrule
        {\makecell[l]{\quad Model \# of Params}} &  \makecell[c]{$124M$} 
        \\
        \midrule
        {\makecell[l]{Generation}}
        \\
        \midrule
        {\makecell[l]{\quad \# of Molecules Optimized}} &  \makecell[c]{$1280$}
        \\
        \midrule
        {\makecell[l]{\quad TopK}} &  \makecell[c]{$[10,15,20]$}
        \\
        \midrule
        {\makecell[l]{\quad TopP}} &  \makecell[c]{$[0.85, 0.9, 0.95]$}
        \\

        \bottomrule
    \end{tabular}}}
        \caption{{{Hyperparameters}}. }
        \label{app:tab:hyperparams}
        }
\end{table*}

\subsection{Results on pretrained REINVENT4}

\begin{table*}[ht!]
\setlength{\tabcolsep}{4pt}
   \centering
    {\small
    \scalebox{0.6}{
    \begin{tabular}{l l c c c c c c c c c c }
        \toprule
        \textbf{Target} %
        & \textbf{Algorithm}
        & {\makecell[c]{Avg \\ Norm Reward~$\uparrow$}}
        & {\makecell[c]{Avg Top 10 \% \\ Norm Reward~$\uparrow$}}
        & {\makecell[c]{Docking ~$\downarrow$}}
        & {\makecell[c]{Druglikeliness ~$\uparrow$}}
        & {\makecell[c]{Synthesizability ~$\downarrow$}}
        & {\makecell[c]{Solubility ~$\uparrow$}}
        & {\makecell[c]{Similarity~$\uparrow$}}
        
        \\
        \midrule
        \makecell[l]{\textbf{3CLPro}} %
        &  \textbf{\makecell[l]{Original}}
        &  \makecell[l]{0.532}
        &  \makecell[l]{{0.689}}
        &  \makecell[l]{-8.698}
        &  \makecell[l]{0.682}
        &  \makecell[l]{3.920}
        &  \makecell[l]{2.471}
        &  \makecell[l]{-}
        \\
        (PDBID:
        &  \textbf{\makecell[l]{MMP \citep{he2022transformer}}}
        &  \makecell[l]{0.629 $\pm$ 0.001}
        &  \makecell[l]{0.717 $\pm$ 0.001}
        &  \makecell[l]{-8.241 $\pm$ 0.015}
        &  \makecell[l]{0.687 $\pm$ 0.003}
        &  \makecell[l]{2.683 $\pm$ 0.005}
        &  \makecell[l]{3.144 $\pm$ 0.028}
        &  \makecell[l]{0.870 $\pm$ 0.003 }
        \\
       \ 7BQY)
        &  \textbf{\makecell[l]{Similarity ($\geq$ 0.5) \citep{he2022transformer}}}
        &  \makecell[l]{0.617 $\pm$ 0.001}
        &  \makecell[l]{0.706 $\pm$ 0.001}
        &  \makecell[l]{-8.222 $\pm$ 0.022}
        &  \makecell[l]{0.690 $\pm$ 0.003}
        &  \makecell[l]{2.664 $\pm$ 0.005}
        &  \makecell[l]{3.162 $\pm$ 0.014}
        &  \makecell[l]{0.803 $\pm$ 0.002 }
        \\
        \textbf{ }
        &  \textbf{\makecell[l]{Similarity ([0.5, 0.7)]) \citep{he2022transformer}}}
        &  \makecell[l]{0.611 $\pm$ 0.001}
        &  \makecell[l]{0.699 $\pm$ 0.001}
        &  \makecell[l]{-8.195 $\pm$ 0.027}
        &  \makecell[l]{0.688 $\pm$ 0.002}
        &  \makecell[l]{\underline{2.660} $\pm$ 0.009}
        &  \makecell[l]{3.196 $\pm$ 0.022}
        &  \makecell[l]{0.775 $\pm$ 0.003 }
        \\
        \textbf{ }
        &  \textbf{\makecell[l]{Similarity ($\geq$ 0.7) \citep{he2022transformer}}}
        &  \makecell[l]{0.630 $\pm$ 0.001}
        &  \makecell[l]{0.717 $\pm$ 0.001}
        &  \makecell[l]{-8.218 $\pm$ 0.007}
        &  \makecell[l]{0.694 $\pm$ 0.001}
        &  \makecell[l]{2.719 $\pm$ 0.006}
        &  \makecell[l]{3.058 $\pm$ 0.021}
        &  \makecell[l]{{0.890} $\pm$ 0.003 }
        \\
        \textbf{ }
        &  \textbf{\makecell[l]{Scaffold \citep{he2022transformer}}}
        &  \makecell[l]{0.607 $\pm$ 0.001}
        &  \makecell[l]{0.704 $\pm$ 0.002}
        &  \makecell[l]{-8.113 $\pm$ 0.015}
        &  \makecell[l]{0.700 $\pm$ 0.002}
        &  \makecell[l]{2.702 $\pm$ 0.006}
        &  \makecell[l]{2.961 $\pm$ 0.014}
        &  \makecell[l]{0.789 $\pm$ 0.002 }
        \\
        \textbf{ }
        &  \textbf{\makecell[l]{Scaffold Generic \citep{he2022transformer}}}
        &  \makecell[l]{0.617 $\pm$ 0.001}
        &  \makecell[l]{0.710 $\pm$ 0.002 }
        &  \makecell[l]{-8.185 $\pm$ 0.017}
        &  \makecell[l]{0.698 $\pm$ 0.002}
        &  \makecell[l]{{2.663} $\pm$ 0.007}
        &  \makecell[l]{3.07 $\pm$ 0.020}
        &  \makecell[l]{0.808 $\pm$ 0.002 }
        \\
        \textbf{ }
        &  \textbf{\makecell[l]{\algname (masks only)}}
        &  \makecell[l]{\underline{0.668} $\pm$ 0.001} 
        &  \makecell[l]{\textbf{0.743} $\pm$ 0.001} 
        &  \makecell[l]{\underline{-9.083} $\pm$ 0.003}
        &  \makecell[l]{\textbf{0.718} $\pm$ 0.001}
        &  \makecell[l]{2.750 $\pm$ 0.001}
        &  \makecell[l]{\underline{3.630} $\pm$ 0.005}
        &  \makecell[l]{0.889 $\pm$ 0.001}
        
        \\
        \textbf{ }
        &  \textbf{\makecell[l]{\algname (mask + s2s)}}
        &  \makecell[l]{\textbf{0.671} $\pm$ 0.001} 
        &  \makecell[l]{\textbf{0.743} $\pm$ 0.001} 
        &  \makecell[l]{\textbf{-9.150} $\pm$ 0.001}
        &  \makecell[l]{\underline{0.714} $\pm$ 0.001}
        &  \makecell[l]{2.763 $\pm$ 0.002}
        &  \makecell[l]{\textbf{3.672} $\pm$ 0.003}
        &  \makecell[l]{{0.895} $\pm$ 0.001}
        \\
        \bottomrule
        \textbf{RTCB}
        &  \textbf{\makecell[l]{Original}}
        &  \makecell[l]{0.536}
        &  \makecell[l]{{0.698}}
        &  \makecell[l]{-8.572}
        &  \makecell[l]{0.709}
        &  \makecell[l]{3.005}
        &  \makecell[l]{2.299}
        &  \makecell[l]{-}
        \\
        (PDBID:
        &  \textbf{\makecell[l]{MMP \citep{he2022transformer}}}
        &  \makecell[l]{0.636 $\pm$ 0.001}
        &  \makecell[l]{0.731 $\pm$ 0.002}
        &  \makecell[l]{-8.422 $\pm$ 0.022}
        &  \makecell[l]{0.712 $\pm$ 0.03}
        &  \makecell[l]{2.601 $\pm$ 0.003}
        &  \makecell[l]{2.987 $\pm$ 0.025}
        &  \makecell[l]{0.851 $\pm$ 0.002 }
        \\
        \ 4DWQ)
        &  \textbf{\makecell[l]{Similarity ($\geq$ 0.5) \citep{he2022transformer}}}
        &  \makecell[l]{0.626 $\pm$ 0.001}
        &  \makecell[l]{0.723 $\pm$ 0.001}
        &  \makecell[l]{-8.452 $\pm$ 0.037}
        &  \makecell[l]{0.712 $\pm$ 0.003}
        &  \makecell[l]{2.579 $\pm$ 0.006}
        &  \makecell[l]{3.013 $\pm$ 0.018}
        &  \makecell[l]{0.785 $\pm$ 0.003}
        \\
        \textbf{ }
        &  \textbf{\makecell[l]{Similarity ([0.5, 0.7)]) \citep{he2022transformer}}}
        &  \makecell[l]{0.622 $\pm$ 0.002}
        &  \makecell[l]{0.718 $\pm$ 0.001}
        &  \makecell[l]{-8.428 $\pm$ 0.016}
        &  \makecell[l]{0.709 $\pm$ 0.002}
        &  \makecell[l]{\underline{2.558} $\pm$ 0.006}
        &  \makecell[l]{3.079 $\pm$ 0.029}
        &  \makecell[l]{0.757 $\pm$ 0.003 }
        \\
        \textbf{ }
        &  \textbf{\makecell[l]{Similarity ($\geq$ 0.7) \citep{he2022transformer}}}
        &  \makecell[l]{0.640 $\pm$ 0.001}
        &  \makecell[l]{0.733 $\pm$ 0.002}
        &  \makecell[l]{-8.445 $\pm$ 0.023}
        &  \makecell[l]{0.718 $\pm$ 0.002}
        &  \makecell[l]{2.629 $\pm$ 0.004}
        &  \makecell[l]{2.880 $\pm$ 0.012}
        &  \makecell[l]{0.880 $\pm$ 0.003}
        \\
        \textbf{ }
        &  \textbf{\makecell[l]{Scaffold \citep{he2022transformer}}}
        &  \makecell[l]{0.615 $\pm$ 0.003}
        &  \makecell[l]{0.720 $\pm$ 0.002}
        &  \makecell[l]{-8.512 $\pm$ 0.038}
        &  \makecell[l]{0.719 $\pm$ 0.001}
        &  \makecell[l]{2.587 $\pm$ 0.005}
        &  \makecell[l]{2.764 $\pm$ 0.014}
        &  \makecell[l]{0.748 $\pm$ 0.002 }
        \\
        \textbf{ }
        &  \textbf{\makecell[l]{Scaffold Generic \citep{he2022transformer}}}
        &  \makecell[l]{0.624 $\pm$ 0.001}
        &  \makecell[l]{0.723 $\pm$ 0.001}
        &  \makecell[l]{-8.497 $\pm$ 0.023}
        &  \makecell[l]{0.722 $\pm$ 0.002}
        &  \makecell[l]{{2.562} $\pm$ 0.006}
        &  \makecell[l]{2.877 $\pm$ 0.019}
        &  \makecell[l]{0.771 $\pm$ 0.002 }
        \\        
        \textbf{ }
        &  \textbf{\makecell[l]{\algname (masks only)}}
        &  \makecell[l]{\underline{0.675} $\pm$ 0.001} 
        &  \makecell[l]{\underline{0.753} $\pm$ 0.001} 
        &  \makecell[l]{\underline{-9.318} $\pm$ 0.002}
        &  \makecell[l]{\textbf{0.752} $\pm$ 0.001}
        &  \makecell[l]{2.674 $\pm$ 0.001}
        &  \makecell[l]{\underline{3.292} $\pm$ 0.002}
        &  \makecell[l]{{0.883} $\pm$ 0.001}
        
        \\
        \textbf{ }
        &  \textbf{\makecell[l]{\algname (mask + s2s)}}
        &  \makecell[l]{\textbf{0.678} $\pm$ 0.001} 
        &  \makecell[l]{\textbf{0.755} $\pm$ 0.001} 
        &  \makecell[l]{\textbf{-9.377} $\pm$ 0.003}
        &  \makecell[l]{\underline{0.751} $\pm$ 0.001}
        &  \makecell[l]{2.688 $\pm$ 0.001}
        &  \makecell[l]{\textbf{3.328} $\pm$ 0.005}
        &  \makecell[l]{{0.890} $\pm$ 0.001}
        \\
        \bottomrule
        \\
    
    \end{tabular}}}
        \caption{
        {\textbf{Main results.} A comparison of seven baselines including Original, six baselines from REINVENT 4 (\emph{without REINFORCE finetuning}) \{MMP, Similarity ($\geq 0.5$), Similarity $\in [0.5,0.7)$, Similarity $\geq 0.7$, Scaffold, Scaffold Generic\} and \algname on multiple objectives 
        based on 3CLPro and RTCB datasets. {The top two results are highlighted as \textbf{1st} and \underline{2nd}.  Results are reported for 5 experimental runs. \algname outperforms the competing baselines in most of the metrics.}
        }
        }
\end{table*}

\clearpage

\end{document}